\documentclass{article}

\usepackage{PRIMEarxiv}

\usepackage[utf8]{inputenc} 
\usepackage[T1]{fontenc}    
\usepackage{hyperref}       
\usepackage{url}            
\usepackage{booktabs}       
\usepackage{amsfonts}       
\usepackage{nicefrac}       
\usepackage{microtype}      
\usepackage{lipsum}
\usepackage{fancyhdr}       
\usepackage{graphicx}       
\graphicspath{{media/}}     
\usepackage{makecell}
\usepackage{bbm}
\usepackage{algpseudocode}

\usepackage{amsmath}
\usepackage{algorithm}

\usepackage{subcaption}

\usepackage{mathtools}

\title{KV-Compress: Paged KV-Cache Compression with Variable Compression Rates per Attention Head
}

\author{
  Isaac Rehg \\
  Cloudflare, inc. \\
  \texttt{irehg@cloudflare.com} \\
}

\begin{document}
\maketitle

\begin{abstract}
Context lengths of Large Language Models (LLMs) have exploded in recent years, with 128k-token context becoming a standard and million-token context becoming a reality. Efficiently supporting long-context inference remains challenging as the memory that must be allocated in key-value (KV) cache for a generation scales with its context length, limiting the number of long-context requests that can be served concurrently under a given memory budget. KV cache compression can mitigate this issue by removing under-utilized KVs from each attention head's cache and reducing its memory footprint. Higher theoretical compression rates can be achieved when the number of removed KVs varies across attention heads \cite{feng2024adakvoptimizingkvcache}, but application of such a strategy within existing inference frameworks adds fragmentation and cannot realize the theoretical compression rates in physical memory. We introduce KV-Compress, a novel compression method that evicts contiguous KV blocks within a PagedAttention \cite{kwon2023efficientmemorymanagementlarge} framework, reducing the memory footprint of the KV cache proportionally to this theoretical compression rate. Our method achieves state-of-the-art performance on LongBench \cite{bai2024longbenchbilingualmultitaskbenchmark} for both Mistral-7B-Instruct-v0.2 and Llama-3.1-8B-Instruct while lowering the total number of compressed KVs by 4x compared with prior methods. Evaluations on Llama-3.1-8B-Instruct and Llama-3.1-70B-Instruct-FP8 achieve compression rates up to 8x with negligible impact on performance, and up to 64x while retaining over 90\% of full-cache performance for all but three of the suite's subsets. We benchmark an integration of our method with vLLM that increases total throughput by up to 5.18x by enabling larger decoding batches \footnote{Code is open-sourced and available at https://github.com/IsaacRe/vllm-kvcompress/tree/main}.
\end{abstract}

\section{Introduction}

Context length has become a key factor in the utility of LLM services with a growing number of providers offering 128k and up context windows. Context windows have been growing in the open-source model ecosystem as well, with the newest round of llama models natively supporting 131k context windows \cite{Meta_2024} and third party finetuning continuing to expand context windows of models after release, even up to 1 million tokens \cite{Feil_2024}. 


Supporting these longer context windows at a large scale is difficult since the number of key and value vectors (KVs) that must be cached when decoding a prompt scales with its total token length. This means that as the context length grows, the maximum batch size that can be used when decoding sequences of this length decreases proportionally. In deployments that are constrained by global GPU memory this can severely limit the system's total throughput, measured in generated tokens per second.

KV cache compression improves the scaling between context-length and KV cache memory by evicting a proportion of less-important KVs from cache. But despite a growing body of research in this space, the integration of these methods into efficient inference platforms such as vLLM and TRT-LLM has been slow.

Most existing methods determine the importance of KVs by comparing their aggregate attention over a number of observed queries. Early approaches use a running aggregate of attention from all past queries \cite{liu2023scissorhandsexploitingpersistenceimportance, zhang2023h2oheavyhitteroracleefficient}, while more recent work has aggregated attention from only the final prompt tokens within a limited \textit{observation window} and seen improvements in overall performance as a result \cite{li2024snapkvllmknowslooking, cai2024pyramidkvdynamickvcache}. We compare the both approaches, controlling for other differences in implementation, and find that aggregating over all past queries outperforms an observation window in many LongBench \cite{bai2024longbenchbilingualmultitaskbenchmark} subsets, despite having a lower average performance rating. Results suggest that there is still room for improvement in state-of-the-art eviction methods by modifying the range of queries over which attention for a given key is aggregated when determining eviction.

Another commonality in prior work is the use of a uniform eviction rate across attention heads of the KV cache. Evicting the same number of KVs ensures that the size of each head's cache remains the same, limiting its fragmentation. Recent work has found that allowing the rate of eviction to vary across attention heads leads to improved performance and enables higher theoretical compression rates \cite{feng2024adakvoptimizingkvcache}. In current inference frameworks, however, a naive application of such an eviction scheme only increases fragmentation and is ineffective at reducing the KV cache memory footprint. We design a modification of PagedAttention \cite{kwon2023efficientmemorymanagementlarge} that can handle the KV cache fragmentation introduced by \textit{variable-head-rate compression} and reduce memory footprint proportionally to the theoretical compression rate.

Building on this work, we introduce KV-Compress, a modification of Ada-SnapKV \cite{feng2024adakvoptimizingkvcache} that evicts KV \textit{blocks}--rather than single KVs--making it compatible with our paged-attention framework. Our method includes further algorithmic improvements including a variable rate of eviction per layer, the use of \textit{squared} past attention to inform evictions, and an extension to grouped-query-attention (GQA) models that handles their cache more effectively. Evaluation on both established benchmarks and state-of-the-art models demonstrates improved performance over existing compression methods.

\begin{figure}
\centering
\begin{subfigure}{.5\textwidth}
  \centering
  \caption{ }
  \vspace{-30pt}
  \label{fig:llama-8b-throughput-acc}
  \includegraphics[width=1.05\linewidth]{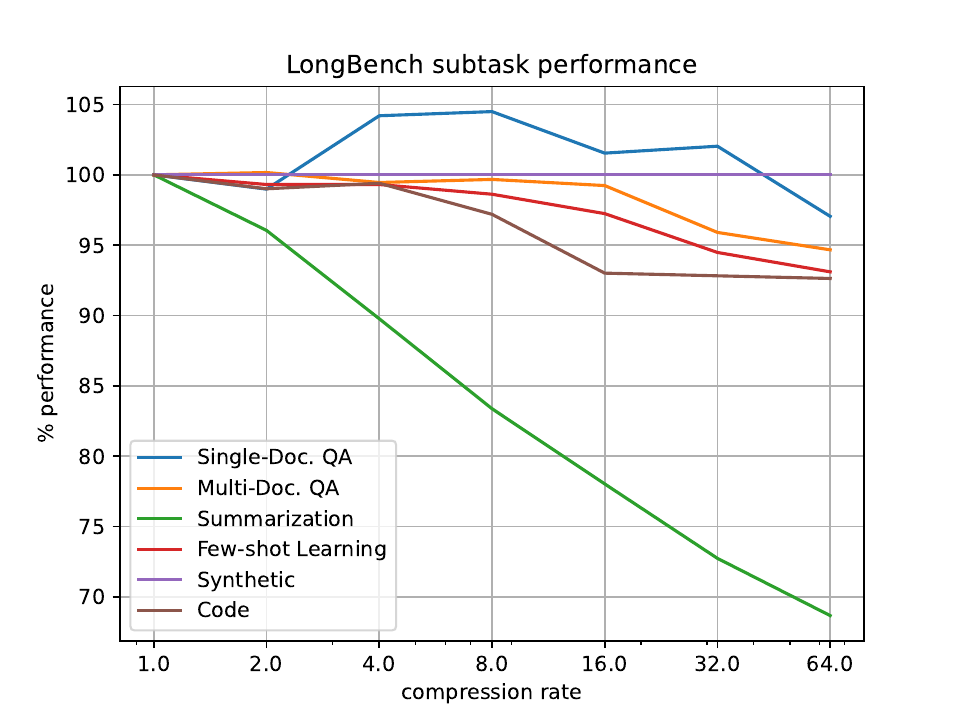}
\end{subfigure}%
\begin{subfigure}{.5\textwidth}
  \centering
  \caption{ }
  \vspace{-30pt}
  \label{fig:llama-8b-throughput-thrpt}
  \includegraphics[width=1.05\linewidth]{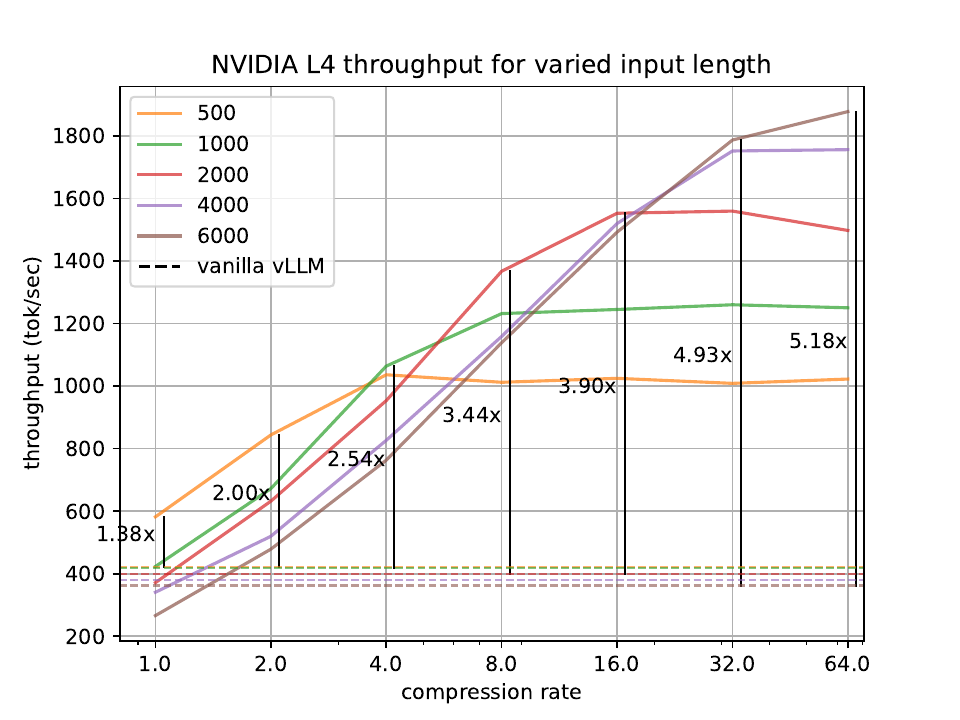}
\end{subfigure}
\caption{Llama-3.1-8B-Instruct performance for different rates of compression. \ref{fig:llama-8b-throughput-acc} plots average LongBench subtask performance for each subtask category, measured as a percentage of the accuracy achieved without compression. \ref{fig:llama-8b-throughput-thrpt} displays the throughput that can be achieved on an NVIDIA L4 for the same range of compression rates. Plots are shown for varying input lengths.}
\end{figure}

Finally, we present an integration of our proposed KV cache compression strategy into vLLM and demonstrate that our method can be used to increase total throughput of state-of-the-art systems for LLM inference several times over when availability of global device memory is a bottleneck.

Our contributions are as follows:
\begin{enumerate}
    \item An analysis of the effect that a limited observation window and its size have on compression performance.
    \item We propose \textit{query-group-compression}, a simple yet effective method to compress the KV cache of GQA models without repeating it into the dimension of total query heads, yielding an additional 4x compression over existing methods on Mistral and Llama-3.1 benchmarks.
    \item An implementation of PagedAttention \cite{kwon2023efficientmemorymanagementlarge} that can represent and compute attention against KV cache containing variable numbers of KVs per head, making variable-head-rate compression practical.
    \item Drawing from these findings we present KV-Compress, a modification of Ada-SnapKV \cite{feng2024adakvoptimizingkvcache} designed to compress a paged KV cache with variable rates of compression across layers and heads. Our method achieves state-of-the-art performance on the LongBench suite \cite{bai2024longbenchbilingualmultitaskbenchmark} for Mistral-7B-Instruct-v0.2 and Llama-3.1-8B-Instruct.
    \item We provide an integration of our method with vLLM and present the first end-to-end benchmarks of an eviction-based KV cache compression method within a paged-attention-enabled framework for efficient LLM inference, to the best of our knowledge. Results demonstrate increased throughput by factors of up to 5.18x.
\end{enumerate}

\begin{table*}[t!]
	\centering
	\addtolength{\tabcolsep}{-4.1pt}
		\renewcommand{\arraystretch}{0.7}
		\begin{tabular}{
				l@{}   c@{\hspace{-0.2ex}}c@{\hspace{0.1ex}}  c@{\hspace{0ex}}c@{\hspace{-2.4 ex}}c@{\hspace{-2.1ex}}c   @{\hspace{-0.7ex}}c@{\hspace{-1.5ex}}c@{\hspace{-1.4ex}}c@{\hspace{0.2ex}}   c@{\hspace{-0.3ex}}c@{\hspace{-1.5ex}} c@{\hspace{-0.3ex}}c@{\hspace{0ex}}c@{\hspace{1.7ex}}c@{\hspace{0.7ex}}c@{\hspace{1.5ex}}c@{}
			}
			\toprule
			& \multicolumn{3}{c}{\small Single-Doc. QA}                                                                                                                                   & \multicolumn{3}{c}{\small Multi-Doc. QA}                                                                                                                                          & \multicolumn{3}{c}{\small Summarization}                                                                                                                                             & \multicolumn{3}{c}{\small Few-shotLearning}                                                                                                                                     & \multicolumn{2}{c}{\small Synthetic}                                                                                & \multicolumn{2}{c}{\small Code}                                                                                   & \multicolumn{1}{c}{}                                  \\ \cmidrule(lr){2-4}\cmidrule(lr){5-7}\cmidrule(lr){8-10}\cmidrule(lr){11-13}\cmidrule(lr){14-15}\cmidrule(lr){16-17}
			& \small \rotatebox[origin=c]{-45}{NrtvQA} & \small \rotatebox[origin=c]{-45}{Qasper} & \small \rotatebox[origin=c]{-45}{MF-en} & \small \rotatebox[origin=c]{-45}{HotpotQA} & \small \rotatebox[origin=c]{-45}{2WikiMQA} & \small \rotatebox[origin=c]{-45}{Musique} & \small \rotatebox[origin=c]{-45}{GovReport} & \small \rotatebox[origin=c]{-45}{QMSum} & \small \rotatebox[origin=c]{-45}{MultiNews} & \small \rotatebox[origin=c]{-45}{TREC} & \small \rotatebox[origin=c]{-45}{TriviaQA} & \small \rotatebox[origin=c]{-45}{SAMSum} & \small \rotatebox[origin=c]{-45}{PCount} & \small \rotatebox[origin=c]{-45}{PRe} & \small \rotatebox[origin=c]{-45}{Lcc} & \multicolumn{1}{l}{\small \rotatebox[origin=c]{-45}{RB-P}} & \small \rotatebox[origin=c]{0}{\makecell{Ave. \\ Score}}  \\ \midrule
			\small Full Cache        & \small 30.49                  & \small 44.83                  & \small 52.88                 & \small 54.96                    & \small 45.70                    & \small 28.41                   & \small 34.41                     & \small 25.63                 & \small 27.05                     & \small 72.50                & \small 91.65                    & \small 43.70                  & \small 6.76                   & \small 97.50               & \small 63.39               & \small 56.73                & \multicolumn{1}{|l}{\small $\:$ 46.30}                     \\ \midrule
			\multicolumn{18}{c}{\small   C=128} \\
			\small H2O          & {\small25.98}        & {\small29.21}                 & {\small41.47}                & {\small50.54}                   & {\small39.35}                   & {\small29.27}                  & {\small23.70}           & {\small22.79}       & {\small22.29}                    & {\small39.00}               & {\small90.48}                   & {\small40.59}        & {\small7.75}                  & {\small99.50}     & {\small56.41}              & \multicolumn{1}{l|}{{\small48.18}}               & {\small$\:$41.66}                \\
			\small SnapKV       & {\small\textbf{29.31}}                 & {\small30.39}                 & {\small48.04}                & {\small51.26}                   & {\small41.35}                   & {\small27.27}                  & {\small22.59}                    & {\small22.82}                & {\small22.33}                    & {\small61.50}               & {\small90.60}                   & {\small40.09}                 & {\small\textbf{8.25}}                  & {\small99.50}              & {\small\textbf{60.89}}              & \multicolumn{1}{l|}{{\small50.80}}               & {\small$\:$44.19}                \\
			\small Pyramid      & {\small28.86}                 & {\small30.72}                 & {\small48.90}                & {\small52.41}                   & {\small39.74}                   & {\small24.36}                  & {\small22.57}                    & {\small22.91}                & {\small22.16}                    & {\small64.50}               & {\small88.65}                   & {\small39.42}                 & {\small7.72}                  & {\small99.50}              & {\small56.73}              & \multicolumn{1}{l|}{{\small49.38}}               & {\small$\:$43.66}                \\
                \midrule
			\small KVC  & {\small29.25}                 & {\small\textbf{33.61}}                 & {\small\textbf{49.72}}                & {\small\textbf{52.64}}                   & {\small\textbf{44.05}}          & {\small\textbf{27.53}}                  & {\small\textbf{23.90}}                    & {\small\textbf{23.48}}                & {\small\textbf{22.91}}           & {\small\textbf{67.00}}      & {\small\textbf{91.47}}                   & {\small\textbf{41.02}}                 & {\small7.00}         & {\small\textbf{100.0}}              & {\small60.82}     & \multicolumn{1}{l|}{{\small\textbf{53.10}}}               & {\small$\:$\textbf{45.47}}       \\ \midrule
			\multicolumn{18}{c}{\small   C=256}  \\
			\small H2O          & {\small28.23}                 & {\small31.68}                 & {\small45.23}                & {\small51.16}                   & {\small41.29}                   & {\small27.90}                  & {\small25.18}           & {\small22.59}                & {\small23.41}                    & {\small39.00}               & {\small90.62}                   & {\small41.51}                 & {\small7.21}                  & {\small99.50}              & {\small59.86}              & \multicolumn{1}{l|}{{\small50.11}}               & {\small$\:$42.78}                \\
			\small SnapKV       & {\small28.66}                 & {\small35.83}                 & {\small50.04}                & {\small52.32}                   & {\small42.02}                   & {\small\textbf{29.84}}                  & {\small24.38}                    & {\small\textbf{23.78}}                & {\small23.71}                    & {\small\textbf{70.00}}               & {\small91.44}                   & {\small41.01}                 & {\small7.27}                  & {\small99.50}              & {\small61.88}              & \multicolumn{1}{l|}{{\small53.20}}               & {\small$\:$45.93}                \\
			\small Pyramid      & {\small28.13}                 & {\small\textbf{38.49}}                 & {\small49.65}                & {\small\textbf{52.99}}                   & {\small\textbf{43.65}}                   & {\small27.36}                  & {\small24.28}                    & {\small23.63}                & {\small23.80}                    & {\small69.50}               & {\small91.14}                   & {\small41.72}                 & {\small\textbf{7.50}}                  & {\small99.50}              & {\small60.47}              & \multicolumn{1}{l|}{{\small53.76}}               & {\small$\:$45.97}                \\
                \midrule
                \small KVC  & {\small\textbf{28.91}}                 & {\small37.56}                 & {\small\textbf{51.73}}                & {\small51.22}                   & {\small43.00}          & {\small28.06}                  & {\small\textbf{25.42}}                    & {\small23.48}                & {\small\textbf{24.14}}           & {\small69.00}      & {\small\textbf{91.86}}                   & {\small\textbf{42.17}}                 & {\small7.40}         & {\small\textbf{99.50}}              & {\small\textbf{62.37}}     & \multicolumn{1}{l|}{{\small\textbf{54.30}}}               & {\small$\:$\textbf{46.26}}       \\ \midrule
			
			\multicolumn{18}{c}{\small   C=512} \\
			\small H2O          & {\small28.48}                 & {\small34.42}                 & {\small46.85}                & {\small50.33}                   & {\small43.56}                   & {\small26.80}                  & {\small26.66}                    & {\small23.08}                & {\small25.04}                    & {\small41.00}               & {\small91.63}                   & {\small41.26}                 & {\small\textbf{7.44}}         & {\small99.50}              & {\small61.57}              & \multicolumn{1}{l|}{{\small51.60}}               & {\small$\:$43.70}                \\
			\small SnapKV       & {\small\textbf{31.20}}        & {\small40.69}                 & {\small51.60}       & {\small53.40}                   & {\small42.63}          & {\small29.33}                  & {\small26.41}                    & {\small23.69}                & {\small25.12}                    & {\small70.50}               & {\small91.73}                   & {\small41.33}                 & {\small7.72}                  & {\small99.50}              & {\small63.62}     & \multicolumn{1}{l|}{{\small55.48}}      & {\small$\:$47.12}                \\
			\small Pyramid      & {\small30.58}                 & {\small40.57}                 & {\small52.54}                & {\small52.57}                   & {\small\textbf{44.03}}                   & {\small27.48}                  & {\small26.52}                    & {\small23.86}                & {\small25.22}                    & {\small70.00}               & {\small\textbf{92.09}}                   & {\small41.70}                 & {\small7.29}                  & {\small99.50}              & {\small62.68}              & \multicolumn{1}{l|}{{\small54.02}}               & {\small$\:$46.92}                \\
                \midrule
                \small KVC  & {\small30.68}                 & {\small\textbf{41.05}}                 & {\small\textbf{53.91}}                & {\small\textbf{54.07}}                   & {\small43.55}          & {\small\textbf{29.50}}                  & {\small\textbf{27.32}}                    & {\small\textbf{23.87}}                & {\small\textbf{25.36}}           & {\small\textbf{71.50}}      & {\small91.86}                   & {\small\textbf{42.94}}                 & {\small7.29}         & {\small\textbf{99.50}}              & {\small\textbf{64.18}}     & \multicolumn{1}{l|}{{\small\textbf{56.51}}}               & {\small$\:$\textbf{47.69}}       \\ \midrule
			\multicolumn{18}{c}{\small   C=1024} \\
			\small H2O          & {\small28.43}                 & {\small38.86}                 & {\small48.95}                & {\small53.38}                   & {\small43.86}                   & {\small26.61}                  & {\small28.49}                    & {\small23.87}                & {\small26.19}                    & {\small45.50}               & {\small91.66}                   & {\small42.32}                 & {\small\textbf{8.11}}         & {\small99.50}              & {\small63.18}              & \multicolumn{1}{l|}{{\small54.50}}               & {\small$\:$45.21}                \\
			\small SnapKV       & {\small30.37}         & {\small44.85}                 & {\small52.20}       & {\small52.87}          & {\small\textbf{44.99}}                   & {\small28.87}                  & {\small\textbf{29.12}}                    & {\small24.33}                & {\small26.40}                    & {\small70.00}               & {\small91.73}          & {\small43.03}                 & {\small7.45}                  & {\small99.50}     & {\small64.11}     & \multicolumn{1}{l|}{{\small56.57}}               & {\small$\:$47.90}                \\
			\small Pyramid      & {\small\textbf{31.04}}                 & {\small\textbf{45.59}}                 & {\small52.93}                & {\small53.52}                   & {\small44.81}                   & {\small\textbf{29.20}}                  & {\small28.67}                    & {\small\textbf{24.42}}       & {\small\textbf{26.45}}                    & {\small70.00}               & {\small91.65}                   & {\small42.74}                 & {\small8.10}                  & {\small99.50}              & {\small\textbf{64.47}}              & \multicolumn{1}{l|}{{\small56.36}}               & {\small$\:$48.09}                \\
                \midrule
                \small KVC  & {\small30.64}                 & {\small43.56}                 & {\small\textbf{53.94}}                & {\small\textbf{54.54}}                   & {\small44.86}          & {\small29.18}                  & {\small29.11}                    & {\small24.10}                & {\small26.28}           & {\small\textbf{72.00}}      & {\small\textbf{91.76}}                   & {\small\textbf{43.04}}                 & {\small7.55}         & {\small\textbf{99.50}}              & {\small64.35}     & \multicolumn{1}{l|}{{\small\textbf{57.95}}}               & {\small$\:$\textbf{48.27}}       \\ \bottomrule
		\end{tabular}
		\caption{Comparison for Llama-3.1-8B-Instruct over 16 LongBench subsets. Highest score for each column shown in bold. For cache size of $C$, all baseline methods keep $C\times 32\times 32$ KVs in cache while our method (KVC) keeps only $C\times 32\times \textbf{8}$ KVs. Table format taken from \cite{feng2024adakvoptimizingkvcache}.}
		\label{tab:detail_llama3}
\end{table*}

\section{Related Work}


\subsection{KV Cache Quantization}
Prior works on KV cache compression attempt to reduce the memory footprint of the KV cache of a transformer language model to enable faster inference over longer contexts. One common approach is KV cache quantization, where compression involves converting all KVs in cache to a reduced-precision representation \cite{zhang2024kvcache1bit, hooper2024kvquant10millioncontext, sheng2023flexgenhighthroughputgenerativeinference}. On the other hand, eviction-based methods reduce cache size by removing KVs that are identified as inconsequential to future decoding. In this work we limit our analysis to methods of the latter category.

\subsection{Token Pruning}
Earlier work has shown that sparsity in the attention mechanism of transformer-based language models can be leveraged to reduce the number of tokens processed across all layers of the model \cite{goyal2020powerbertacceleratingbertinference}. The tokens selected for pruning should be inconsequential--meaning the degree of attention to their keys is low--so that their removal does not adversely affect the output. 

One approach is to prune tokens in a layer-wise manner during the initial processing of an input context \cite{Wang_2021}. At each layer, attention allocated to input tokens is summed over all attention heads and the generated KVs for the tokens with lowest total attention are removed. This has the advantage of speeding up input processing as well as decoding, but has the potential for error to compound as pruning decisions made at earlier layers do not account for the attention in the later layers they are affecting. 

\subsection{KV Eviction}
Since the advent of LLMs and long-context there has been increased interest in methods that prune KVs from cache \textit{after} input processing, or "prefill", with a focus on more efficient decoding. Evicting KVs from cache lowers its memory footprint, enabling the support of larger batches/context windows, and can speed up decoding as the number of KVs to retrieve when computing attention is reduced.

Approaches vary in where/when KV pruning is applied, but most rely on an aggregation of attention that a particular key--or group of keys--has received over past queries. This metric is used when considering KVs for removal from cache, with KVs who receive the least attention being removed first.

\textbf{Full Observation Window}
H2O \cite{zhang2023h2oheavyhitteroracleefficient} determines evictions based on a running sum of attention each KV has received, evicting KVs with the lowest total attention and retaining "heavy-hitter" KVs that receive the most attention. In this case the token positions of evicted KVs are allowed to vary, but the total number of evicted KVs across all layers and heads is fixed.
Scissorhands \cite{liu2023scissorhandsexploitingpersistenceimportance} follow a similar approach, but evict based on a metric of how "pivotal" each KV is--framed as the number of the occurrences where attention with that key has exceeded some threshold. Like H2O, they evict the same number of KVs across attention heads, but vary eviction rate across layers proportionally to a "persistence ratio" determined by an initial profiling phase.
FastGen \cite{ge2024modeltellsdiscardadaptive} combines the "heavy-hitter" approach of H2O with heuristic eviction policies that retain only KVs for special tokens, punctuation or local attention, profiling attention over the input prompt to determine which strategy to employ. 

Though decoding-time efficiency can be improved by these techniques, the preliminary task of aggregating attention of each key over all queries can make prefill prohibitively expensive as doing so requires writing the full attention matrix to global memory. Efficient attention implementations such as Flash Attention \cite{dao2022flashattentionfastmemoryefficientexact} explicitly avoid this in order to speed up inference and reduce global memory requirements. Because of this, performance gains made at decoding time from employing such approaches are unlikely to outweigh the performance loss experienced during prefill, especially for long contexts.

\textbf{Limited Observation Window}
SnapKV \cite{li2024snapkvllmknowslooking} mitigates this issue by limiting the aggregation to queries generated for the final tokens of the input prompt. This limited observation window reduces scaling of the eviction metric calculation from  $O(L^2)$ to $O(L)$ and improves accuracy of the compression. To improve accuracy of eviction metrics obtained from their limited observation window the authors apply maxpooling over the sequence dimension, so that keys neighboring a heavy-hitter of the same attention head are retained as well. This approach works well for prompts consisting of long context followed by a request or question since the attention patterns observed over the prompt queries--occurring within the observation window--are generally similar to those exhibited by queries in the generated response. Like H2O, they allow different eviction schedules to be specified per layer and head, but require that the number of KVs evicted for each layer and head be the same.


Following this work, PyramidKV \cite{cai2024pyramidkvdynamickvcache}, adopts the limited observation window of SnapKV but configures variable rates of eviction per layer. The authors recognize that attention is more evenly distributed at early layers, motivating them to configure lower eviction rates at early layers and evicting more aggressively at later layers. The exact eviction rate is determined initially for the first and last layers during a profiling phase, and eviction rates for remaining layers are inferred using a novel interpolation technique.

\textbf{Variable-Head-Rate Eviction}
More recent work has identified a key point of rigidity in prior methods--namely, a uniform compression rate across all the heads within each layer \cite{nawrot2024dynamicmemorycompressionretrofitting, feng2024adakvoptimizingkvcache}. One such approach develops variations of both SnapKV and PyramidKV that evict KVs \textit{across} heads in each layer, potentially evicting a variable number of KVs per head, naming the new methods Ada-SnapKV and Ada-PyramidKV, respectively \cite{feng2024adakvoptimizingkvcache}. They find that this modification yields improved performance--measured over 16 subsets of LongBench \cite{bai2024longbenchbilingualmultitaskbenchmark}--for both approaches. Their reported results remain theoretical, however, since existing inference frameworks cannot efficiently represent a KV cache with variable number of KVs per head without suffering from memory fragmentation. As a result, evicting along one head without evicting along all other heads only adds fragmentation to the allocated tensor without reducing its footprint in device memory. Our approach uses a modification of PagedAttention to represent and compute attention over a paged KV cache where the fragmentation added by variable-head-rate eviction is reduced, allowing us to realize the theoretical compression rate in physical memory.

Dynamic Memory Compression \cite{nawrot2024dynamicmemorycompressionretrofitting} introduces a learned approach where the base model is trained to either add each new KV to cache or "accumulate" it into an existing KV slot via a weighted average, where decisions are made separately across heads. At inference time their approach uses PagedAttention to efficiently facilitate attention over variably compressed attention heads.

Related to these approaches, ThinK \cite{xu2024thinkthinnerkeycache} was developed to prune the channels of remaining KVs after other eviction-based compression methods have been carried out. The method achieves an additional 30-40\% compression before observing further performance degradation.

\begin{figure}
\centering
\begin{subfigure}{.5\textwidth}
  \centering
  \caption{ }
  \vspace{-30pt}
  \includegraphics[width=1.05\linewidth]{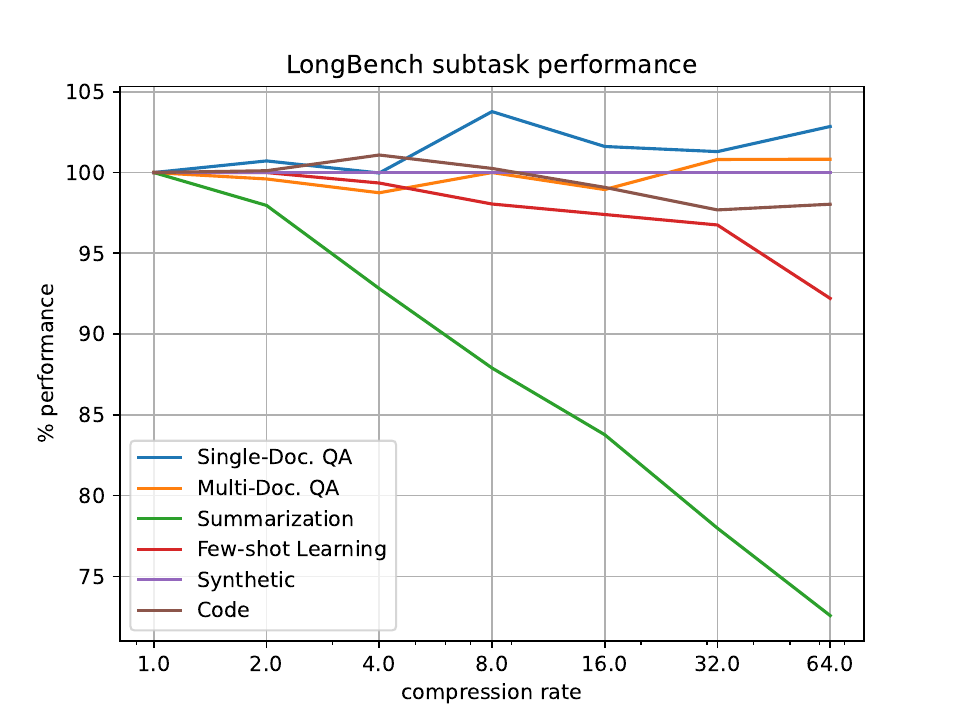}
  \label{fig:llama-70b-throughput-acc}
\end{subfigure}%
\begin{subfigure}{.5\textwidth}
  \centering
  \caption{ }
  \vspace{-30pt}
  \includegraphics[width=1.05\linewidth]{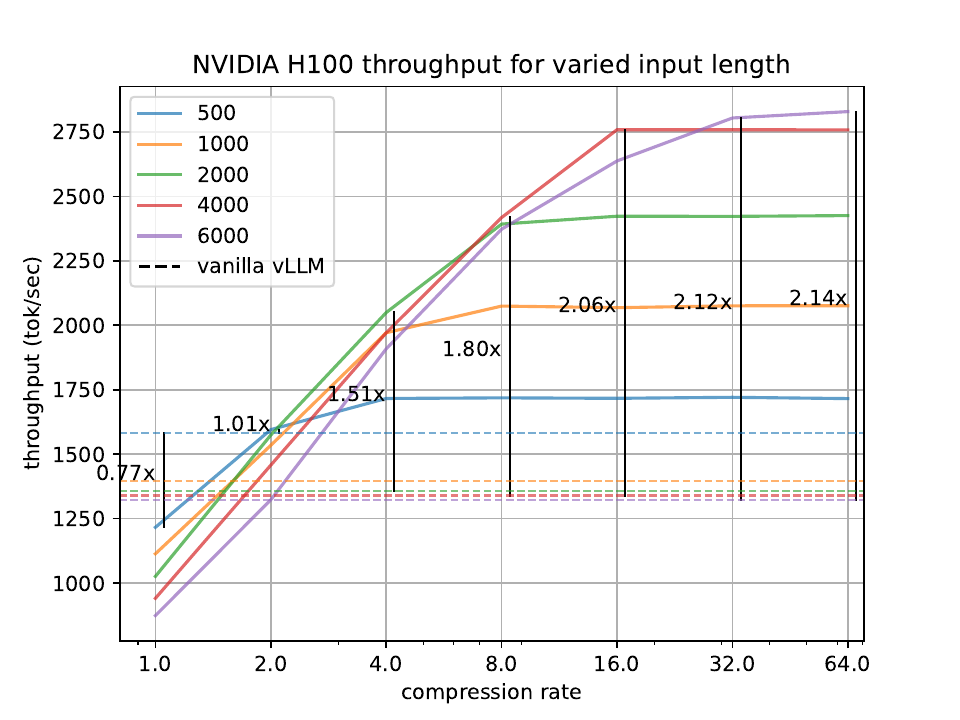}
  \label{fig:llama-70b-throughput-thrpt}
\end{subfigure}
\caption{Llama-3.1-70B-Instruct-FP8 performance for different rates of compression. \ref{fig:llama-70b-throughput-acc} plots average LongBench subtask performance for each subtask category, measured as a percentage of the accuracy achieved without compression. \ref{fig:llama-70b-throughput-thrpt} displays the throughput that can be achieved on an NVIDIA H100 for the same range of compression rates. Plots are shown for varying input lengths.}
\label{fig:test}
\end{figure}
\section{Preliminaries}
The methods outlined in this paper build off of prior work around KV cache compression and their application is made possible through a modification of PagedAttention \cite{kwon2023efficientmemorymanagementlarge}. We will briefly go over these topics before diving into our methodology.

\subsection{Multi-head Attention}
Transformer decoders use attention layers to build a contextualized representation of an input sequence before sampling the next token. Attention operates on sequences of query, key and value vectors, $Q$, $K$ and $V$. The attention matrix for a sequence is computed as

\begin{equation}
    A = \text{softmax}\left(QK^T/\sqrt{d}\right)\;
    \quad\text{for }\; Q_{L\times d} \;, 
    \;K_{L\times d} \;, \;V_{L\times d}
\end{equation}
where $L$ is the sequence length and $d$ is the magnitude of the $q$, $k$, $v$ vectors. The attention output is then computed as the inner product of the attention matrix and $V$. Causal masking is applied before the softmax, so that
\begin{equation}
    \left(QK^T/\sqrt{d}\right)_{ij} = -\inf\;, \quad \forall i < j\;,
\end{equation}

making $A$ lower-triangular after applying the softmax.

In practice, most transformer models employ \textit{multi-head attention} (MHA), where attention occurs in parallel over a series of $H$ attention "heads". In this case, a separate set of queries, keys and values are computed for each head so that $A$ has shape $H \times L \times L$.

\subsection{Paged Attention}

\begin{figure}
    \centering
    \hspace*{-12pt}
    \includegraphics[width=1.03\linewidth]{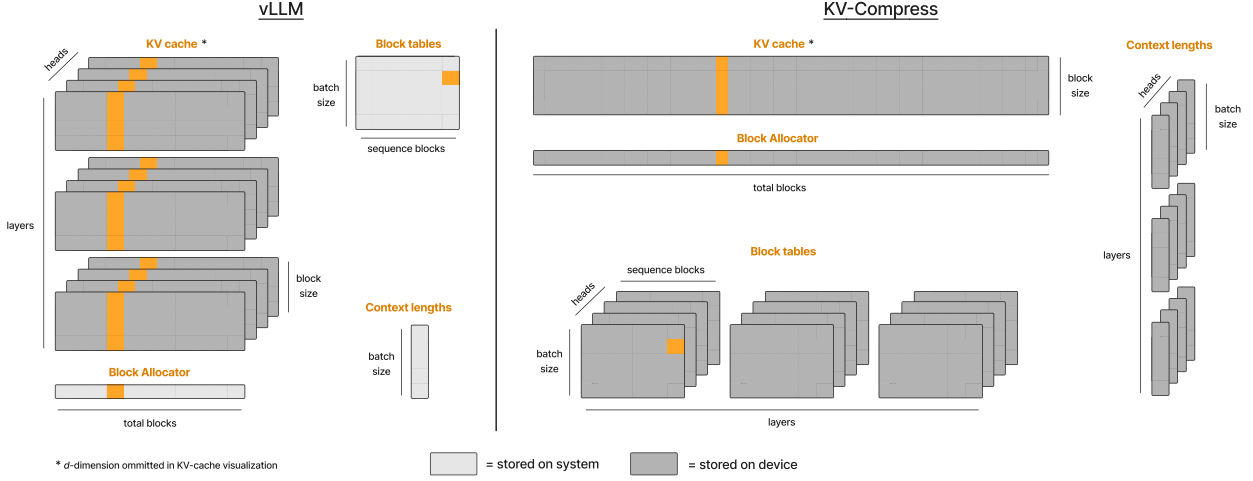}
    \caption{Comparison of block management in vanilla vLLM versus with KV-Compress. In vLLM (left) each cache block stores KVs for every attention head of every layer, while with KV-Compress (right) each block only holds KVs for a single head. The KV-Compress block tables are therefore expanded by $l \times H$ so that the unique block for each specific KV head and layer can be retrieved. Due to the larger number of blocks, we move block management to the GPU so scheduling operations can be done in parallel. Note that every cell of KV cache in the above diagram represents a $d$-dimensional vector, while all others represent scalars.}
    \label{fig:block-mngmt}
\end{figure}

When conducting LLM inference, key vectors and value vectors (KVs) are generated for every attention head at each layer to be used when computing the attention output for all future tokens of that layer and head. During decoding it is common to cache these KVs to avoid recomputing them during each decoding step. In the simplest case, the key and value cache for each layer takes the shape
\( B \times H \times L \times d \) where \(B\) is the decoding batch size, \(H\) is the number of attention heads, \(L\) is the current max sequence length and \(d\) is the magnitude of key and value vectors. In this case total memory allocation scales with the maximum sequence length in the decoding batch and the more that sequence length varies across batch elements, the more fragmentation is introduced and the less efficient global memory utilization is.

PagedAttention \cite{kwon2023efficientmemorymanagementlarge} solves this problem by introducing a paged KV cache where physical memory is allocated in \textit{blocks} of size $b$, such that each block contains KVs across all heads for $b$ tokens of a particular sequence and we only allocate KVs blocks for each sequence's currently generated KVs. This approach is used inference frameworks such as vLLM, and is depicted in figure \ref{fig:block-mngmt} (left).

PagedAttention uses a block table, $T$ with shape

\begin{align}
    B \times \frac{L_{\max}}{b} \;,
\end{align}

where $L_{\max}$ is the maximum allowed sequence length, to index into the corresponding KV cache block for each block of a given sequence. Because each token generates a single key and value vector for every attention head of every layer, the layout of KVs is identical across all layers and heads. PagedAttention, therefore, uses a shared index to reference a particular token's KVs across all layers and heads of the paged KV cache, so that a single cache block contains the set of KVs generated across all layers and heads for each token in that block. Each block then contains $l \times H \times b$ key and value vectors, where $l$ is the number of layers in the model. The physical key cache, $K$, and value cache, $V$ consist of $l$ tensors, each with shape $N \times H \times b \times d$, where $N$ is the total number of allocable cache blocks.

To retrieve key and value vectors, $k$ and $v$ for layer $m$ and head $h$ of a token in sequence $s$ at position $i$ in the paged KV cache, you first retrieve the block number for the physical cache block containing that token's KVs. This can be done by computing
\begin{align}
    n = T_{s,u}\;\quad\text{for}\;
    u = \left\lfloor \frac{i}{b}\right\rfloor\;.
\end{align}

We then index into our physical cache with

\begin{align}
    k = K_{n,h,o}^{(m)}\;,\quad
    v = V_{n,h,o}^{(m)}\quad\text{for}\quad o = i\;\text{mod}\;b
\end{align}

where $K^{(m)}$ and $V^{(m)}$ are the physical key and value cache of layer $m$, respectively and $o$ is the positional offset of token $i$ within block $n$. With this approach the number of allocated but unused KV slots is at most $lBH (b - 1)$ and variance in the context lengths of sequences in the decoding batch does not increase cache fragmentation.

Since the memory allocation for every sample in a batch is no longer fixed, batch size can be scaled dynamically based on the token length of batch elements. To handle this dynamic scaling, frameworks will generally perform an initial scheduling step to determine whether new input prompts can be prefilled and added to the decoding batch or whether current batch elements need to be preempted to make room for new decoded tokens. In vLLM, during the scheduling step a \textit{block manager} allocates available blocks of cache to store new tokens that will be generated during the next prefill or decoding step. If there are no available blocks, one or more sequences will be selected for preemption and the block manager will free their allocated blocks. The block manager tracks running context length of all sequences in the decoding batch and manages a list of free and allocated blocks to determine scheduling and preemption of sequences during each scheduling step. Because block tables generally reside in system memory, scheduling can be CPU intensive for large batch sizes.

\subsection{KV Cache Compression}
KV cache compression has been explored as another means to handling KV cache memory allocation more efficiently, by evicting KVs in cache that are predicted to receive low attention in the future and can therefore be removed with negligible impact on the output. By removing a constant number of KVs from each attention head of a given layer's KV cache the \(L\)-dimension of the KV cache tensor for that layer can be decreased and total memory allocation reduced.

For a particular compression scheme, its compression rate is commonly defined as the ratio of KVs in cache before compression to KVs in cache after compression. In past work, the number of KVs in cache is usually fixed during experimentation using a configured number of \textit{maximum cache tokens}, $C$. In such cases the number of KVs in cache is limited to the number of KVs that would be generated by the model if a prompt of length $C$ were to be processed and cached without compression.

To determine which KVs to evict, a metric is employed to predict the effect that removing a given KV will have on the inference given the attention observed between that key and queries within some range. A commonly used eviction metric is a sum over attention of past queries to each key. H2O and SnapKV both take this approach, with H2O aggregating attention over all observed queries and SnapKV aggregating attention over only the last \(w\) prompt queries.

We can define the metrics obtained from aggregating over attention of all queries as

\begin{equation}
    \label{full-window}
    M_{hj}^{(full)} = \sum_i{A_{hij}}
\end{equation}

and can define the metrics obtained from aggregating over a limited observation window spanning the last $w$ prompt queries as

\begin{equation}
    \label{observation-window}
    M_{hj}^{(w)} = \sum_{i=s}^{L}{A_{hij}} \;
    ,\quad\text{for}\; s  = L - w\;.
\end{equation}

Keys within the observation window are safe from eviction since they will have fewer attention observations to sum over due to the causal masking of $A$.

SnapKV applies additional max-pooling over $L$ to improve performance, computing metrics as

\begin{align}
    \label{pooled-metrics}
    M_{h,j}^{(pool)} = \max_{t=j-p/2}^{j+p/2}{M_{h,t}^{(w)}}\;
\end{align}

for pooling size $p$.

Running the compression for a particular layer involves sorting $M$ along the sequence length dimension and then selecting KVs corresponding to the first $e$ metrics along each attention head for eviction. The compressed cache can then be obtained by concatenating the remaining key and value vectors along each attention head, yielding a tensor of shape $H \times (L - e) \times d$.

\subsection{Grouped-Query Attention}
Most modern LLM architectures employ grouped-query-attention (GQA) \cite{ainslie2023gqatraininggeneralizedmultiquery} as a means of speeding up inference and reducing the size of KV cache. GQA works by reducing the number of key-value heads, $n_k$ to a fraction $n_k = \frac{n_q}{r}$ of the number of query heads, $n_q$ and evenly dividing the query heads into $n_k$ groups with $r$ heads, each. Each group of query heads is assigned a single key-value head that produces key and value vectors for every query head in the group during the multi-head attention operation. Naive implementations of GQA accomplish this by performing an interleaved repeat of the generated KV tensors over the attention head dimension before conducting scaled-dot-product attention. Efficient inference frameworks, on the other hand, generally rely on specialized kernels such as those of FlashAttention \cite{dao2022flashattentionfastmemoryefficientexact} or PagedAttention \cite{kwon2023efficientmemorymanagementlarge} that can lookup the proper key and value vector for queries of a given query group without explicitly allocating space for repeated KVs. Applying GQA with such an implementation can, in fact, be seen as a form of KV cache compression where the compression rate is equal to $r$, the ratio of query heads to key-value heads, since the GQA model will cache $\frac{1}{r}$ as many KVs during inference as a comparable MHA model.

\begin{figure}
\centering
\includegraphics[width=1\linewidth]{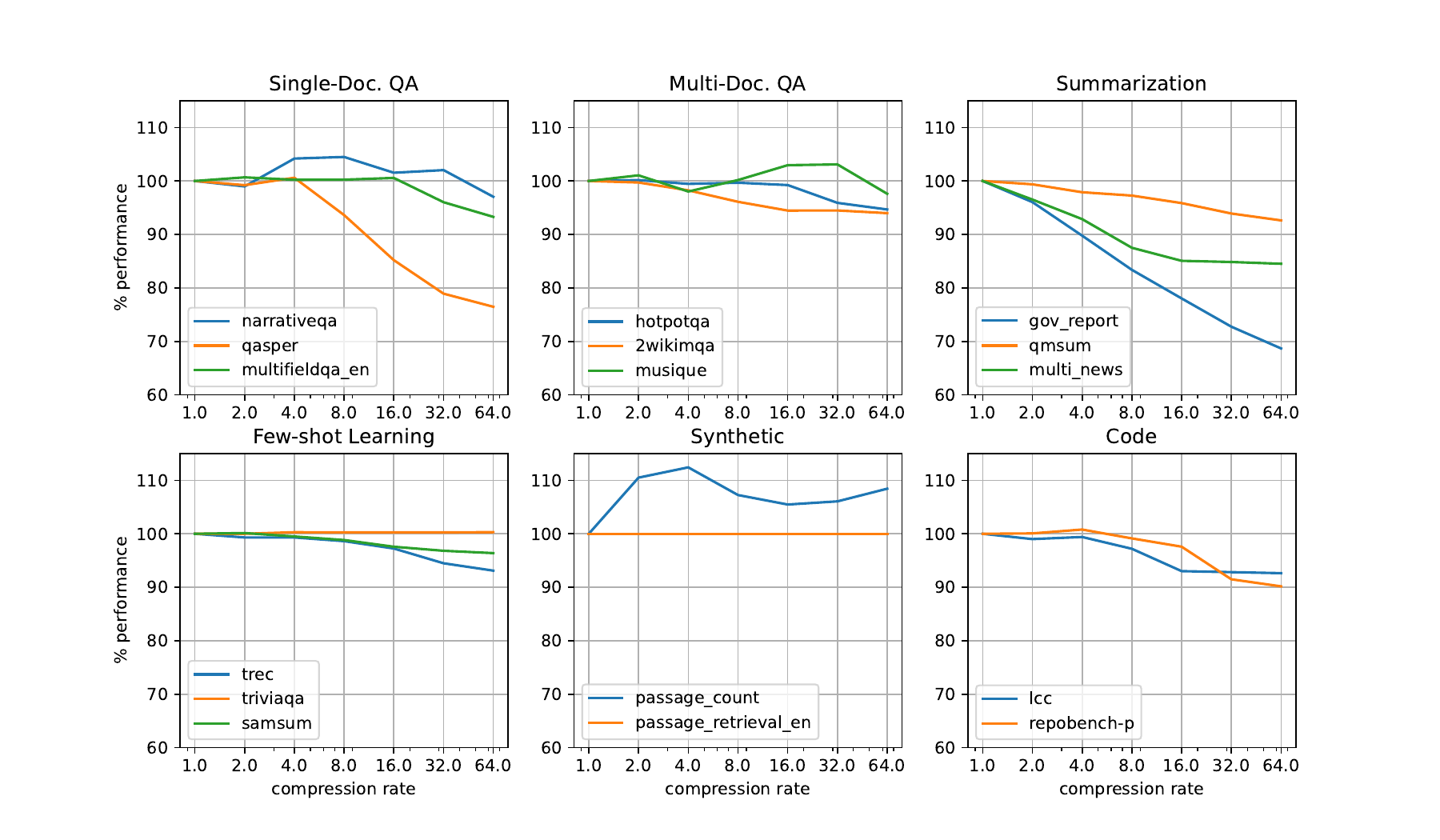}
\caption{Llama-3.1-8B-Instruct percent of full-cache performance by compression rate on all LongBench subtasks. Results are grouped by subtask category.}
\label{fig:llama-8b-longbench-all}
\end{figure}

\section{Method}
The method presented in this paper is a modification of Ada-SnapKV \cite{feng2024adakvoptimizingkvcache}. We first discuss query-group compression, then detail our adaptation of PagedAttention that makes variable-head-rate eviction practical. Finally we go over the remaining algorithm choices made in our final method.


\subsection{Query Group Compression}
While GQA has seen widespread adoption over the last year in models such as Llama and Mistral, existing methods for KV cache compression were not designed with a GQA KV cache in mind. Implementations of both SnapKV and PyramidKV, for example, cache and compress KV tensors \textit{after} repetition is carried out for alignment with the query tensor. This means that: 1. The compression rate must exceed the ratio of query heads to KV heads in order to improve upon the compression that would be achieved by using a more efficient framework where KVs are not physically repeated in memory. 2. There is a great deal of redundancy in cache before compression occurs (in the case of Mistral and Llama-3, $\frac{3}{4}$ of KVs are duplicates) and this redundancy is not being taken advantage of in the compression.

We seek a compression method where KVs are evicted from a non-repeated cache that is applicable to current GQA models run in state-of-the-art inference frameworks. This can be done with a straightforward modification to existing eviction-based methods, where the metrics used to determine KV eviction are aggregated for each key over queries in that key's respective query group. We can then continue with compression of the non-repeated cache, using the aggregate metric to inform eviction decisions.

Following this modification, equation \ref{full-window} becomes

\begin{align}
    \label{full-window-gqa}
    M_{h_k,j}^{(full)} = \sum_i\;\sum_{h\in H_k}{A_{hij}} \;
    ,\quad\text{for}\; H_k = \left\{\; \forall\; h: rh_k \leq h < r(h_k+1) \;\right\}
\end{align}

where we add an additional summation over the metrics computed for all queries in the current key's query group, $H_k$. Similarly, equation \ref{observation-window} becomes

\begin{align}
    M_{h_k,j}^{(w)} = \sum_{i=s}^{L}\;\sum_{h\in H_k}{A_{hij}} \;
    ,\quad\text{for}\; s  &= L - w \\
    H_k &= \left\{ \;\forall\; h: rh_k \leq h < r(h_k+1) \;\right\} \;.
\end{align}

\subsection{Supporting Variable-Head-Rate Eviction}

Ada-SnapKV explores evicting a variable number of KVs from each attention head of the KV cache. Selecting KVs to evict in this case can be done by sorting metrics over a flattened tensor where head and sequence 
 length dimensions are combined, then selecting KVs corresponding to the first $eH$ metrics for eviction. Unlike Ada-SnapKV, we seek to additionally support variable rates of compression across \textit{layers}, following this same methodology. In this section we discuss the steps taken to make such compression feasible.

\subsubsection{Block Layout and Parallel Allocation}
The application of variable-head-rate eviction within existing inference frameworks is ineffective as it only reduces the cache size proportionally to the attention head with \textit{lowest} compression rate and all evictions beyond this rate merely increase cache fragmentation. We note this is conceptually similar to how sequences of variable length in a decoding batch create fragmentation in a non-paged KV cache, as the cache size scales with maximum sequence length.

To reconcile with this added fragmentation we can adapt PagedAttention to page out cache on a per-head, per-layer--as well as per sequence--basis. We expand the block tables of each sequence to include block tables for each layer and attention head of the cache, so that they can be retrieved for each KV head during attention without the use of fixed memory offsets.

This gives us block tables of shape

\begin{align}
    B \times l \times H \times \frac{L_{\max}}{b}\;,
\end{align}
compared to the layout used in vanilla PagedAttention, $B \times \frac{L_{\max}}{b}$.

Each block now only contains KVs for a single KV head of a single layer. We utilize a single, contiguous physical cache allocation for all layers, of shape $N \times b \times d$, compared to the $l$ layer-specific physical cache allocations of shape $N \times H \times b \times d$ used by most paged-attention frameworks.

Since the scheduling step in paged-attention frameworks happens off-device, its runtime scales linearly as the number of blocks being allocated increases. In the KV-Compress block layout this presents an issue, as the number of blocks is $lH$ times larger than that of standard frameworks. With a naive extension of vLLM's scheduler, we observe prohibitively expensive scheduling loops--in some cases taking longer to complete than the model forward pass. To remedy this we design an on-device block allocation system where both block tables and context lengths for each attention head of each sequence's cache are stored on device. This allows us to parallelize both the counting of allocated blocks and the allocation of new blocks.

When scheduling for prefill we can compute the necessary blocks from the token length, alone, since each sequence will initially allocate the same number of blocks per head. When scheduling decoding of running sequences we compute the required additional block allocations for all sequences in parallel from the on-device context lengths tensor. When preempting, we compute the number of freed blocks across all layers and heads in parallel, for each preempted sequence. We use a flat on-device tensor of length $N$ to track allocation of blocks in the unified KV cache. Before the forward pass, block tables and context lengths are passed directly from the block manager to the model runner. Our block management layout is visualized in figure \ref{fig:block-mngmt} (right).

\subsubsection{Evicting from a Paged KV Cache}

\begin{figure}
    \centering
    \hspace*{-20pt}
    \includegraphics[width=1.05\linewidth]{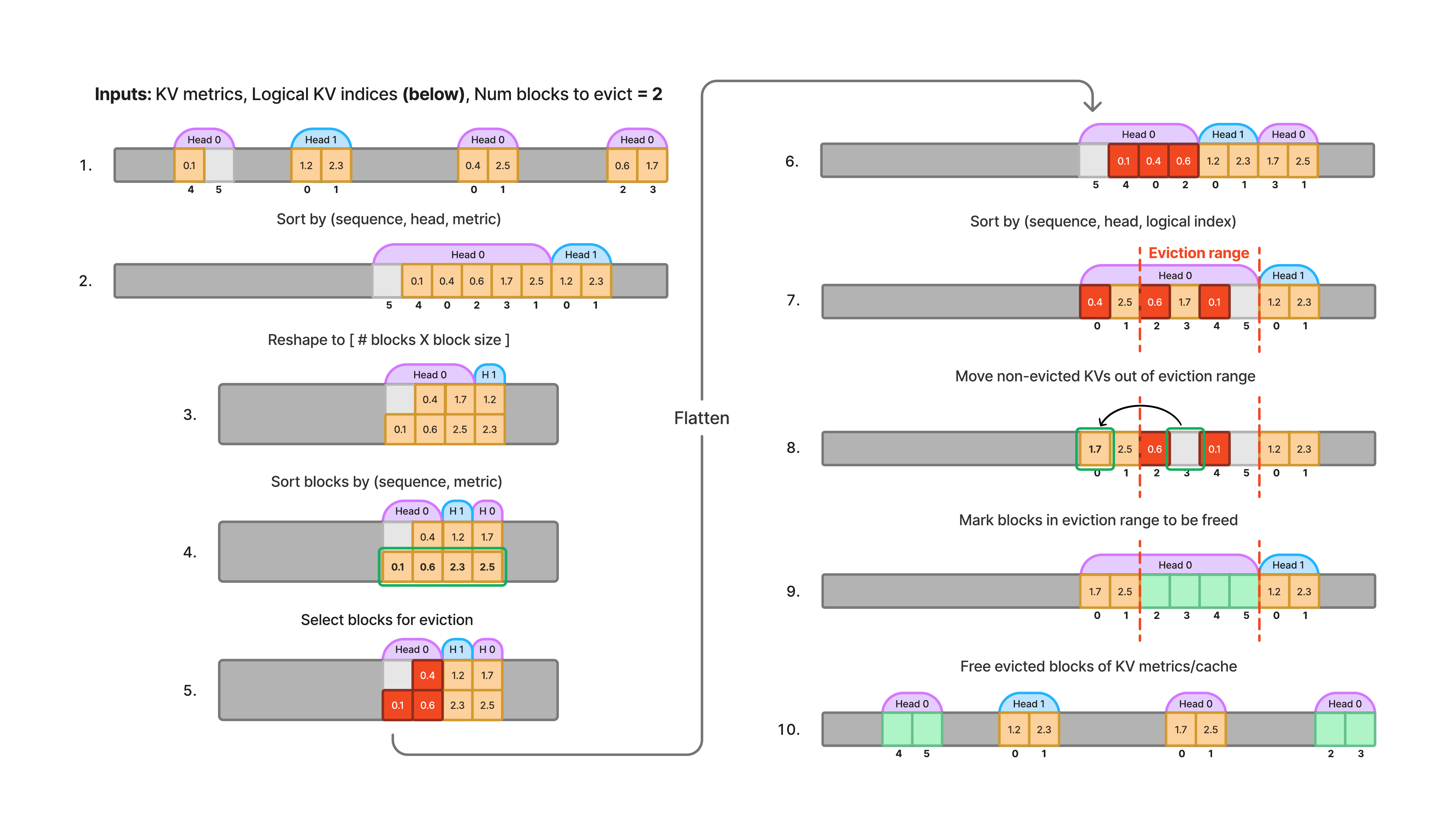}
    \caption{Visualization of our compression algorithm for a simplified example with two KV heads and a block size of two. KV metrics are visualized for a given cache state, highlighting blocks of a particular sequence in the decoding batch that is scheduled to evict two blocks. Logical indices are displayed under the corresponding metrics slot.}
    \label{fig:compression-alg}
\end{figure}

To support KV cache compression within a paged-attention framework, we need to be able to evict \textit{blocks} rather than \textit{KVs}, as this is what determines total memory footprint of the KV cache. We seek an algorithm to evict entire blocks of KVs such that the sum eviction metric over all KVs in the evicted cache block of a particular head is equivalent to the sum eviction metric that would result from evicting $b$ KVs from that head in a fully unconstrained manner.

More concretely, for a KV head, $h$ with $N$ allocated KV blocks, we want an ordering, $S_h$, of sets of $b$ KVs along $h$, satisfying

\begin{alignat}{2}
    S_h = (K^{(1)}, K^{(2)}, ..., K^{(N)})\quad
    \text{s.t.}\quad\;&\left|K^{(i)}\right| = b\;,&&\quad\forall i \\
    &K^{(i)} \cap K^{(j)} = \emptyset\;,&&\quad\forall i \neq j \\
    &\sum_{f\in K^{(i)}}{m_{f}} \leq \sum_{g\in K^{(j)}}{m_{g}}
    \;,&&\quad \forall i < j\;
\end{alignat}

where $m_i$ denotes the eviction metric for a KV, $i$.

By setting a small enough block size, $b$, we can then achieve comparable performance to the more flexible approach where eviction counts are not limited to multiples of $b$.

To satisfy the requirements for $S_h$, KVs will need to be rearranged in physical cache, as it is improbable that the next-$b$ most suitable KVs for eviction will all occur within the same block initially. Our algorithm for determining block eviction and carrying out this reorganization is displayed in figure \ref{fig:compression-alg}. As we walk through the algorithm we will use the superscript, $^{(t)}$, to denote variables at each step, $t$.

We start with our unified physical key and value caches, $K_u$ and $V_u$, both of shape $N \times b \times d$ and our metrics tensor, $M$ of shape $N \times b$, following the same layout as $K_u$ and $V_u$ along dimensions $N$ and $b$. We also need the ordering of KVs along each head, represented as a vector of size $N$ in the same layout as $M$, mapping KVs of each head to their position in this ordering. We call this the \textit{logical index} of a KV, and though it is initially defined as the token position at which the KV was generated, this will change once we rearrange the cache during compression.

Because the metrics and KV cache follow a row-major layout in memory, we can view them as a two-dimension $Nb \times d$ tensor and one-dimensional $Nb$ tensor, respectively, where the metrics and KVs within the same block are adjacent in memory.

Figure \ref{fig:compression-alg}.1 displays an example metrics tensor, $M^{(1)}$, following this view, with metrics for a particular sequence highlighted. This simplified example visualizes compression of a single layer model with only two KV heads. The sequence has been previously compressed and has five and two KVs along its first and second KV heads, respectively. Logical indices are displayed below each metric slot. The physical KV cache (not shown) follows the same layout, with each metric corresponding to a previously allocated key or value vector.

Our goal is to minimize the metrics of evicted KVs, so we first need to identify, for each head, $h$, the the maximum-valued metric that will be evicted when evicting $e$ blocks from that head's cache, $m(h, e)$. Let $C_h$ be the the number of KVs allocated for $h$, so that the number of allocated blocks is given by $\left\lceil\frac{C_h}{b}\right\rceil$. We have

\begin{align}
    \text{dom}(m) = \left\{ \forall h, e: 1 \leq e < \left\lceil\frac{C_h}{b}\right\rceil \right\}\;.
\end{align}

Since we are free to rearrange KVs in cache, $m(h, e)$ is simply the $(be)^{\text{th}}$ smallest metric in the cache of head $h$. We can compute $m(h, e)$ over its full domain by first sorting $M^{(1)}$ by $(\text{head}, \text{metric})$ to get $M^{(2)}$, then reshaping $M_{Nb \times 1}^{(2)}$ into $M_{N \times b}^{(3)}$ to get

\begin{align}
    m(h, e) = M_{o_h+e-1,b-1}^{(3)}
\end{align}

where $o_h$ is the offset in physical blocks to the first metric block for head $h$. The empty slot in the last allocated block is treated as having a metric of zero, so that it comes before all non-empty metrics of its head when sorted. This ensures that the maximum metric for the first evicted block of each head is only aggregated over the smallest $a$ metrics, where $a=C_h\; \text{mod}\; b\;$ is the number of KVs that reside in the last allocated block and must be evicted before that head's first whole cache block can be freed.

The columns of $M^{(3)}$ now define per-head KV eviction schedules, as each $M_{i,:}^{(3)}$ contains the next $b$ lowest metrics for corresponding head, $h$, when $i-o_h$ of its blocks have already been evicted. Next, we use $m(h, e)$ for each block of eviction candidates to inform our eviction rates across heads. This is done by sorting $M_{N \times b}^{(3)}$ along the $N$-dimension by the values, $M_{:,b-1}^{(3)}\;$, to get $M^{(4)}$. This preserves the block-contiguity of $M^{(3)}$ and gives us an ordering of candidate block evictions by lowest max metric across heads. It's this layout that lets us obtain our final KV eviction schedule for a particular sequence, given its budget of total retained cache blocks for that compression iteration. 

For each sequence, $s$ we take its budgeted number of total block evictions, $E_s$ and mark all KVs in the first $E_s$ blocks for eviction, computing the eviction mask

\begin{align}
    W_{i,:}^{(5)} = \sum_s\;\mathbbm{1}(O_s \leq i < O_s + E_s)
\end{align}

where $O_s$ is the offset in physical blocks to the first metric block for sequence $s$.

In the example our sequence is forced to evict two cache blocks, and the most suitable block evictions as determined by our algorithm both come from the first KV head. Had the sequence been forced to evict an additional block it would evict the lone block of its second head.

Once we have this mask we can reshape it into $W_{Nb \times 1}^{(6)}$ and sort by logical index to get $W^{(7)}\;$, which follows the original layout of $M^{(2)}$. Note that in this layout the KVs that were scheduled for eviction are no longer contiguous in memory, and their eviction alone will not enable the freeing of any cache blocks since every block in our example still contains at least one non-evicted KV. To resolve this issue we need to reorganize our KVs in physical memory so that the eviction status of KVs in each block is homogeneous. We know that such a reordering is possible, as scheduling was conducted such that the sum count of evicted KVs and empty cache slots is evenly divisible by the block size.


\begin{algorithm}
  \caption{MoveCache algorithm for reorganizing evicted KVs into contiguous blocks\\
  \textbf{Input:} K cache $K$, V cache $V$, metrics $\textbf{m}$, logical indices $\textbf{p}$, eviction mask $\textbf{w}$, evicted block count $e$, block size $b$}\label{alg:movecache}
  \begin{algorithmic}[1]
    \Procedure{MoveCache}{$K,V,\textbf{m},\textbf{p},\textbf{w},e,b$}
      \State $i\gets 0$
      \State $j\gets |\textbf{p}|-1$
      \State $end\gets j-eb$\Comment{Define the eviction range}
      \While{j > end}
        \While{$\textbf{w}_i = 0$}\Comment{Find next evicted K outside eviction range}
            \State $i\gets i + 1$
        \EndWhile
        \If{$\textbf{w}_j = 0$}\Comment{We have a non-evicted K inside eviction range}
            \State $src\gets \textbf{p}_j$
            \State $dst\gets \textbf{p}_i$
            \State $K_{dst,:}\gets K_{src,:}$\Comment{Move KV vectors}
            \State $V_{dst,:}\gets V_{src,:}$
            \State $\textbf{m}_{dst}\gets \textbf{m}_{src}$\Comment{Move corresponding metrics}
        \EndIf
        \State $j\gets j-1$
      \EndWhile
    \EndProcedure
  \end{algorithmic}
\end{algorithm}

Our \textit{MoveCache} algorithm, outlined in algorithm \ref{alg:movecache} and depicted in figure \ref{fig:compression-alg}.8, resolves this problem by defining an eviction range spanning the last $E_s$ blocks of a sequence, then iterating backwards over KVs wintin this range by decreasing logical index, moving any non-evicted KVs it encounters in place of the next evicted KV occurring outside the eviction range. Once iteration over the eviction range has completed, reorganization is complete, and we can map blocks within the eviction range of the reordered metrics back to the corresponding block in the original physical cache layout, to be freed and made available for future allocation.

\subsubsection{Limiting Overhead}
The main source of overhead in our compression algorithm comes from sorting over KV metrics. We use PyTorch's \textit{sort} API to  compute each compression step where a reordering is required. These calls add overhead in the form of an increased memory footprint and added latency during compression. From profiling on an NVIDIA L4, we observe an additional memory allocation of around 8 times the size of the sorted tensor, and find that runtime begins to scale linearly when size exceeds $1.7e8$. To both minimize the memory that needs to be reserved for the sort operations and prevent them from exhausting GPU resources we limit the number of KVs that can be compressed during a given compression iteration, as each KV corresponds to an element in the metrics tensor that will need sorting.

\begin{figure}
\centering
\includegraphics[width=1\linewidth]{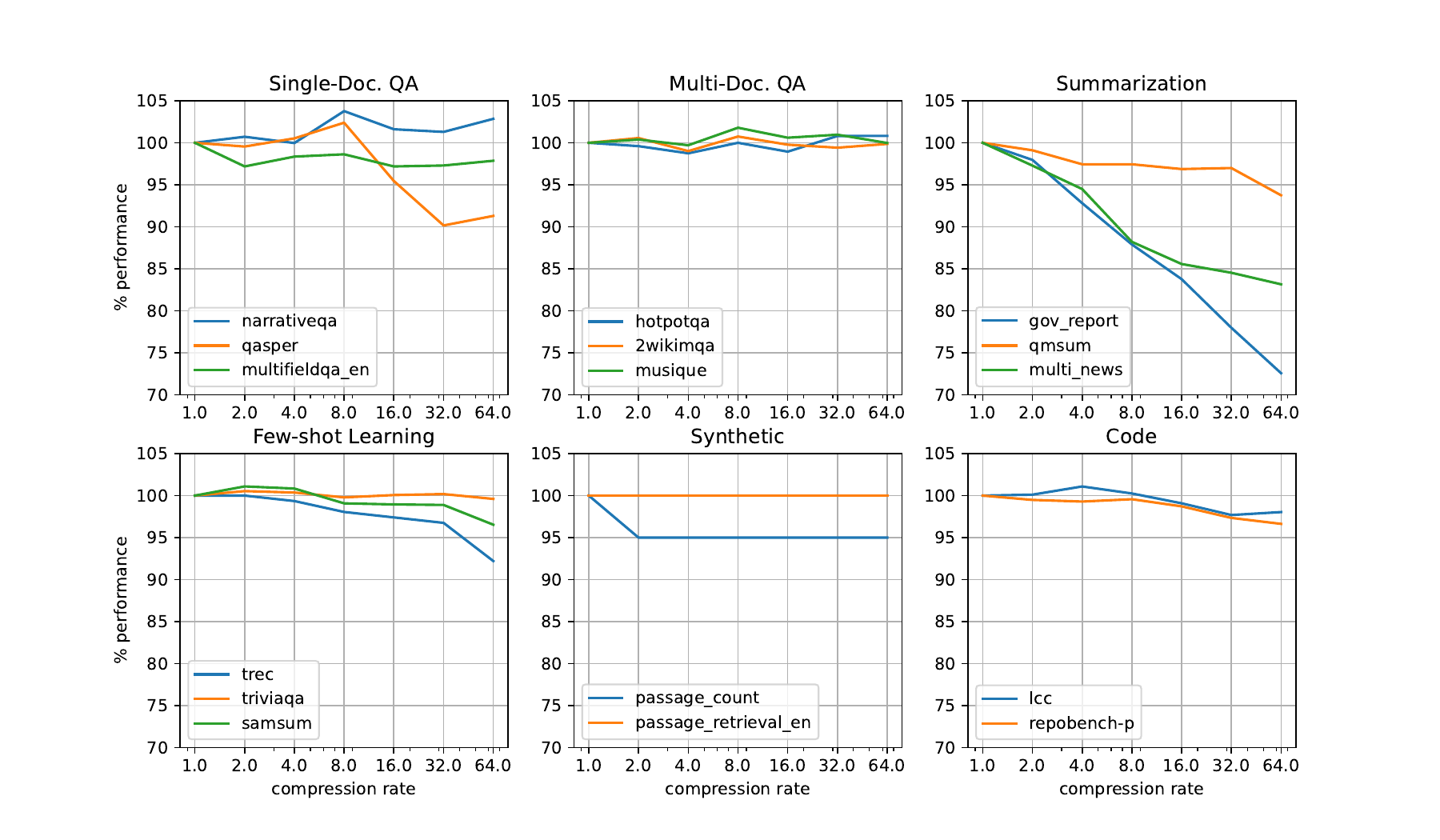}
\caption{Llama-3.1-70B-Instruct-FP8 percent of full-cache performance by compression rate on all LongBench subtasks. Results are grouped by subtask category}
\label{fig:llama-70b-longbench-all}
\end{figure}

Whenever a round of compression is initiated, we iterate over the current set of running sequences, ordered by time of last compression (with uncompressed sequences coming first), 
adding to the compression batch and tracking the total number of KVs currently allocated to sequences in this batch as we go. When we encounter a sequence that would cause the batch's total allocated KVs to exceed our configured limit, we discard it and move forward with compression of the current batch of selected sequences.

Setting this limit too low can prevent sequences from being scheduled for compression if their total number of allocated KVs exceeds it. A simple solution is to configure the limit to be at least equal to $L_{\max} \times l \times H$, where $L_{\max}$ is the maximum supported input token length. This ensures that the total number of KVs generated for a sequence at any given point stay below the compression limit, so that its compression is guaranteed once all other sequences have been compressed.

To further reduce overhead we seek to limit the number of compression iterations that occur, relative to the number of model forward passes. We identify several approaches to controlling the scheduling of compression steps:
\begin{enumerate}
    \item Compress every $c$ model iterations, for some fixed compression interval, $c$.
    \item Compress whenever the count of total uncompressed tokens passes some threshold.
    \item Compress when one or more sequences are newly prefilled.
    \label{compress-on-prefill}
    \item Compress when we would otherwise be forced to preempt a running sequence.
    \label{compress-on-preempt}
\end{enumerate}

After experimenting with these approaches, we find the combination of \ref{compress-on-prefill} and \ref{compress-on-preempt} to be the most effective, and use this in our final method.

\subsection{Metric Calculation}
In this section we discuss in detail our method for computing the metrics for each KV that are used to determine our eviction schedule.

\textbf{Squared Attention Metric}
Most prior work sum attention values between each key and queries within the observation window to get each KVs eviction metric. In this case the eviction schedule can be thought of as attempting to minimize L1 error in future attention. Alternatively, we can seek to minimize the \textit{L2} error over future attention by using a sum of squared attention to compute eviction metrics. We experiment with both approaches and find that squared attention perform better, so we take this route in our approach.

\textbf{Observation Window}
In prior work, aggregating attention to compute KV metrics for eviction has had two approaches: Aggregate over all past observed KVs, as in H2O; or aggregate over a partial window of queries as in SnapKV. To explore the tradeoffs of these two approaches, we design two distinct versions of our method.

\textit{KVC-full} aggregates squared attention over all observed queries. We find it beneficial to exclude queries of the first \(v\) tokens following that KV's token from the aggregation, as the local attention patterns that occur between keys and queries within close proximity of one another are generally not representative of the attention patterns that will occur between the same key and future queries to that are generated outside of that key's immediate locality. We adapt equation \ref{full-window-gqa} to define our metrics as

\begin{align}
    M_{h_k,j}^{(full)} = \sum_{i=j+v}^L\;\sum_{h\in H_k}{(A_{hij})^2} \;
    ,\quad\text{for}\; H_k = \left\{ \;\forall \;h: rh_k \leq h < r(h_k+1) \;\right\}
\end{align}

for an excluded query window, $v$.


\textit{KVC-w} aggregates squared attention over queries within a limited observation window of final tokens of the input prompt. Metrics are computed following equation \ref{pooled-metrics}, with the slight modification of squaring the attention values before summation.

\textbf{Continual Compression}
Our compression method can be easily adapted to conduct KV cache compression during decoding, either during regular intervals of decoding steps or on an as-needed basis. When compression is continued during decoding, we accumulate squared attention from queries of newly generated tokens directly into the KV metrics computed during prefill. In this case the metric for a KV generated at position $j$ at decoding step \(t\) is given by

\begin{align}
    M_{k_h,j}^{(cc)} = M_{h_k,j}^{(pool)} + \sum_{i=L_c}^{L_c+t}\;\sum_{h\in H_k}{(A_{hij})^2} \;
    ,\quad\text{for}\; H_k = \left\{ \;\forall \;h: rh_k \leq h < r(h_k+1) \;\right\}
\end{align}

where $L_c$ is the length of input context and $M^{(pool)}$ is the set of metrics computed during prefill.

\section{Experiments}

\subsection{Baselines}

We evaluate performance of KV-Compress on the same 16 subsets of the LongBench test suite used in prior work \cite{li2024snapkvllmknowslooking, cai2024pyramidkvdynamickvcache, feng2024adakvoptimizingkvcache}. Following prior work we measure performance for varying levels of maximum cache size, $C=\{128, 256, 512, 1024\}$, evicting KVs such that the total number of KVs over all layers and heads is equivalent to the number of KVs that would be generated by an input with token length $C$.

We evaluate our compression on both Mistral-7B-Instruct-v0.2 and Llama-3.1-8B-Instruct, comparing performance against the following baseline methods introduced in prior work:
\begin{itemize}
    \item H2O \cite{zhang2023h2oheavyhitteroracleefficient}
    \item SnapKV \cite{li2024snapkvllmknowslooking}
    \item PyramidKV \cite{cai2024pyramidkvdynamickvcache}
    \item Ada-SnapKV and Ada-PyramidKV \cite{feng2024adakvoptimizingkvcache}
\end{itemize}

For Mistral baselines, we take the results published by \cite{feng2024adakvoptimizingkvcache} after spot-checking to verify reproducibility.
For Llama baselines, we use the implementation of \cite{cai2024pyramidkvdynamickvcache} to run and evaluate all methods, adapting their code to support transformers version 4.44.2 for compatibility with the rope-scaling configuration of Llama-3.1 \footnote{Code is available at https://github.com/IsaacRe/PyramidKV}.

For both Llama and Mistral, we evaluate KV-Compress model outputs using the same methods followed in \cite{li2024snapkvllmknowslooking} and \cite{cai2024pyramidkvdynamickvcache}.

Because no implementation for Ada-SnapKV/PyramidKV was available at the time of release, we compare only against their published results on Mistral-7B-Instruct-v0.2.

All baselines aside from H2O employ an observation window when collecting eviction metrics. We follow \cite{li2024snapkvllmknowslooking}
in setting observation window size to $w=8$ and pooling size to $p=7$ for all baseline evaluations.

We evaluate KV-Compress variations KVC-full and KVC-w against the above baselines. We denote the type of aggregation used to obtain KV metrics for KV-Compress by appending a suffix to the method (\textit{-L2} for squared sum and \textit{-L1} for standard sum). If not explicitly specified, KV-Compress uses $w=8$ and L2 aggregation.

\subsection{Settings}
We run all KV-Compress experiments using our vLLM integration forked from v0.6.0, running in eager mode with a block size of 16. For all L4/Llama-8B experiments, 
we use default gpu memory utilization of $0.9$ and set max-model-length to 19,000. For all H100/Llama-70B experiments, we set gpu memory utilization to 0.96 and limit max-model-length to 33,000 to account for global memory limitations. For all other configurations we use default run parameters of v0.6.0. 

KVC-full uses a excluded query window of $v=10$ in all experiments and KVC-w uses a pooling size of $p=7$. When running \textit{KVC-full} on an H100 we find it necessary to limit gpu memory utilization to 0.6 to reserve space for the expensive metric collection. To limit required memory for metric collection we compute the aggregate over one query block at a time, using a block size of 1024.

In all baseline comparisons we run a single iteration of compression following prefill, using the specified max-cache-size, $C$.

For throughput benchmarks, we use vLLM's benchmarking script, modified to configure compression rate and other operational parameters. For these experiments, max-cache-size is configured per-sample as $\min(128, \frac{1}{r}L_c)$, for compression rate $r$ and input context length $L_c$. We schedule iterations of compression after each prefill and whenever preemption would otherwise be forced.
We compare against baseline performance of a clean install of vLLM v0.6.0 on equivalent hardware, keeping all parameters unrelated to compression the same. Each benchmarking run is conducted over 256 input prompts. We vary the input prompt length per-run, but keep the number of output tokens fixed at 500.
Each data point for Llama-3.1-8B is averaged over three separate runs. Experiments for Llama-3.1-70B were conducted with a single run due to compute limitations.

\subsection{LongBench}

\begin{table*}[t!]
	\centering
	\addtolength{\tabcolsep}{-5.4pt}
		\renewcommand{\arraystretch}{0.7}
		\begin{tabular}{
				l@{}   c@{\hspace{-0.2ex}}c@{\hspace{0.1ex}}  c@{\hspace{0ex}}c@{\hspace{-2.4 ex}}c@{\hspace{-2.1ex}}c   @{\hspace{-0.7ex}}c@{\hspace{-1.5ex}}c@{\hspace{-1.4ex}}c@{\hspace{0.2ex}}   c@{\hspace{-0.3ex}}c@{\hspace{-1.5ex}} c@{\hspace{-0.3ex}}c@{\hspace{0ex}}c@{\hspace{1.7ex}}c@{\hspace{0.7ex}}c@{\hspace{1.5ex}}c@{}
			}
			\toprule
			& \multicolumn{3}{c}{\small Single-Doc. QA}                                                                                                                                   & \multicolumn{3}{c}{\small Multi-Doc. QA}                                                                                                                                          & \multicolumn{3}{c}{\small Summarization}                                                                                                                                             & \multicolumn{3}{c}{\small Few-shotLearning}                                                                                                                                     & \multicolumn{2}{c}{\small Synthetic}                                                                                & \multicolumn{2}{c}{\small Code}                                                                                   & \multicolumn{1}{c}{}                                  \\ \cmidrule(lr){2-4}\cmidrule(lr){5-7}\cmidrule(lr){8-10}\cmidrule(lr){11-13}\cmidrule(lr){14-15}\cmidrule(lr){16-17}
			& \small \rotatebox[origin=c]{-45}{NrtvQA} & \small \rotatebox[origin=c]{-45}{Qasper} & \small \rotatebox[origin=c]{-45}{MF-en} & \small \rotatebox[origin=c]{-45}{HotpotQA} & \small \rotatebox[origin=c]{-45}{2WikiMQA} & \small \rotatebox[origin=c]{-45}{Musique} & \small \rotatebox[origin=c]{-45}{GovReport} & \small \rotatebox[origin=c]{-45}{QMSum} & \small \rotatebox[origin=c]{-45}{MultiNews} & \small \rotatebox[origin=c]{-45}{TREC} & \small \rotatebox[origin=c]{-45}{TriviaQA} & \small \rotatebox[origin=c]{-45}{SAMSum} & \small \rotatebox[origin=c]{-45}{PCount} & \small \rotatebox[origin=c]{-45}{PRe} & \small \rotatebox[origin=c]{-45}{Lcc} & \multicolumn{1}{l}{\small \rotatebox[origin=c]{-45}{RB-P}} & \small \rotatebox[origin=c]{0}{\makecell{Ave. \\ Score}}  \\ \midrule
			\small Full Cache        & \small 26.63                  & \small 32.99                  & \small 49.34                 & \small 42.77                    & \small 27.35                    & \small 18.77                   & \small 32.87                     & \small 24.24                 & \small 27.10                     & \small 71.00                & \small 86.23                    & \small 42.96                  & \small 2.75                   & \small 86.98               & \small 55.33               & \small 52.87                & \multicolumn{1}{|l}{\small $\:$ 42.51}                     \\ \midrule
			\multicolumn{18}{c}{\small   C=128} \\
			\small H2O          & {\small21.19}        & {\small21.66}                 & {\small38.60}                & {\small30.63}                   & {\small20.65}                   & {\small12.19}                  & {\small20.65}           & {\small22.42}       & {\small21.81}                    & {\small39.00}               & {\small82.52}                   & {\small\textbf{40.68}}        & {\small2.98}                  & {\small\textbf{79.56}}     & {\small49.13}              & \multicolumn{1}{l|}{{\small46.76}}               & {\small$\:$34.40}                \\
			\small SnapKV       & {\small19.17}                 & {\small21.40}                 & {\small42.93}                & {\small36.76}                   & {\small22.44}                   & {\small15.86}                  & {\small19.16}                    & {\small21.84}                & {\small21.55}                    & {\small47.50}               & {\small84.15}                   & {\small40.24}                 & {\small2.30}                  & {\small68.26}              & {\small50.69}              & \multicolumn{1}{l|}{{\small47.13}}               & {\small$\:$35.09}                \\
			\small Pyramid      & {\small20.16}                 & {\small21.77}                 & {\small43.55}                & {\small36.78}                   & {\small23.12}                   & {\small14.39}                  & {\small19.53}                    & {\small22.03}                & {\small21.47}                    & {\small51.00}               & {\small84.62}                   & {\small40.24}                 & {\small2.79}                  & {\small70.77}              & {\small50.57}              & \multicolumn{1}{l|}{{\small46.53}}               & {\small$\:$35.58}                \\
			\small Ada-SKV   & {\small20.63}                 & {\small22.58}        & {\small45.68}       & {\small37.90}          & {\small23.49}                   & {\small16.55}         & {\small19.99}                    & {\small22.28}                & {\small21.55}                    & {\small59.50}               & {\small85.00}          & {\small40.62}                 & {\small3.09}                  & {\small69.36}              & {\small50.98}              & \multicolumn{1}{l|}{{\small48.17}}      & {\small$\:$36.71}                \\
			\small Ada-PKV  & {\small20.50}                 & {\small21.71}                 & {\small45.61}                & {\small36.81}                   & {\small23.57}          & {\small15.84}                  & {\small19.75}                    & {\small22.13}                & {\small22.00}           & {\small60.50}      & {\small84.04}                   & {\small40.51}                 & {\small3.21}         & {\small73.60}              & {\small51.24}     & \multicolumn{1}{l|}{{\small48.02}}               & {\small$\:$36.81}       \\ \midrule
			\small w32-L1  & {\small20.15}                 & {\small20.30}                 & {\small45.32}                & {\small37.70}                   & {\small24.29}          & {\small15.20}                  & {\small19.92}                    & {\small21.98}                & {\small21.36}           & {\small51.75}      & {\small84.72}                   & {\small39.96}                 & {\small3.46}         & {\small68.05}              & {\small51.55}     & \multicolumn{1}{l|}{{\small47.93}}               & {\small$\:$35.85}       \\ 
			\small w32-L2  & {\small19.74}                 & {\small22.79}                 & {\small46.27}                & {\small38.31}                   & {\small\textbf{24.78}}          & {\small15.40}                  & {\small20.52}                    & {\small22.77}                & {\small21.62}           & {\small63.25}      & {\small\textbf{85.31}}                   & {\small40.16}                 & {\small\textbf{3.50}}         & {\small70.78}              & {\small52.43}     & \multicolumn{1}{l|}{{\small49.61}}               & {\small$\:$37.33}       \\ 
			\small w8-L2  & {\small\textbf{22.58}}                 & {\small25.04}                 & {\small\textbf{48.11}}                & {\small35.50}                   & {\small24.01}          & {\small14.44}                  & {\small22.19}                    & {\small\textbf{22.84}}                & {\small21.98}           & {\small68.50}      & {\small84.28}                   & {\small39.87}                 & {\small3.43}         & {\small66.81}              & {\small\textbf{53.02}}     & \multicolumn{1}{l|}{{\small\textbf{49.70}}}               & {\small$\:$\textbf{37.64}}       \\ 
			\small full-L2  & {\small22.20}                 & {\small\textbf{25.34}}                 & {\small42.45}                & {\small\textbf{40.16}}                   & {\small24.54}          & {\small\textbf{17.62}}                  & {\small\textbf{25.40}}                    & {\small22.05}                & {\small\textbf{25.18}}           & {\small\textbf{70.50}}      & {\small84.87}                   & {\small35.36}                 & {\small3.33}         & {\small44.05}              & {\small50.78}     & \multicolumn{1}{l|}{{\small45.41}}               & {\small$\:$36.20}       \\ \midrule
			\multicolumn{18}{c}{\small   C=256}  \\
			\small H2O          & {\small21.54}                 & {\small22.92}                 & {\small42.56}                & {\small31.07}                   & {\small22.53}                   & {\small13.76}                  & {\small22.52}           & {\small22.40}                & {\small23.09}                    & {\small40.50}               & {\small84.20}                   & {\small40.77}                 & {\small3.41}                  & {\small86.10}              & {\small50.98}              & \multicolumn{1}{l|}{{\small48.17}}               & {\small$\:$36.03}                \\
			\small SnapKV       & {\small22.37}                 & {\small23.74}                 & {\small48.13}                & {\small38.56}                   & {\small22.43}                   & {\small15.66}                  & {\small21.91}                    & {\small23.13}                & {\small23.15}                    & {\small61.50}               & {\small85.45}                   & {\small41.42}                 & {\small3.09}                  & {\small84.54}              & {\small53.22}              & \multicolumn{1}{l|}{{\small50.24}}               & {\small$\:$38.66}                \\
			\small Pyramid      & {\small20.09}                 & {\small24.00}                 & {\small47.33}                & {\small38.24}                   & {\small22.48}                   & {\small16.02}                  & {\small21.40}                    & {\small22.45}                & {\small22.63}                    & {\small63.00}               & {\small84.93}                   & {\small40.98}                 & {\small3.40}                  & {\small82.48}              & {\small52.78}              & \multicolumn{1}{l|}{{\small49.36}}               & {\small$\:$38.22}                \\
			\small Ada-SKV   & {\small22.55}                 & {\small25.78}        & {\small48.33}       & {\small40.30}          & {\small24.24}          & {\small16.64}                  & {\small21.63}                    & {\small23.03}                & {\small23.19}           & {\small67.00}      & {\small85.78}          & {\small41.53}        & {\small\textbf{3.47}}         & {\small\textbf{87.07}}     & {\small53.86}     & \multicolumn{1}{l|}{{\small51.13}}      & {\small$\:$39.72}       \\
			\small Ada-PKV  & {\small22.64}        & {\small24.64}                 & {\small47.40}                & {\small40.25}                   & {\small23.62}                   & {\small16.83}         & {\small21.82}                    & {\small23.34}       & {\small22.70}                    & {\small66.50}               & {\small84.99}                   & {\small41.34}                 & {\small2.78}                  & {\small86.90}              & {\small53.17}              & \multicolumn{1}{l|}{{\small49.52}}               & {\small$\:$39.28}                \\ \midrule
                \small w32-L1  & {\small22.05}                 & {\small24.45}                 & {\small48.98}                & {\small38.47}                   & {\small24.36}          & {\small16.16}                  & {\small21.94}                    & {\small22.99}                & {\small23.10}           & {\small62.75}      & {\small85.58}                   & {\small41.37}                 & {\small2.83}         & {\small79.99}              & {\small53.05}     & \multicolumn{1}{l|}{{\small51.43}}               & {\small$\:$38.72}       \\ 
                \small w32-L2  & {\small22.82}                 & {\small26.41}                 & {\small48.28}                & {\small39.85}                   & {\small\textbf{25.30}}          & {\small16.39}                  & {\small22.74}                    & {\small23.29}                & {\small23.33}           & {\small67.25}      & {\small85.32}                   & {\small41.63}                 & {\small2.97}         & {\small85.52}              & {\small53.86}     & \multicolumn{1}{l|}{{\small51.80}}               & {\small$\:$39.80}       \\ 
                \small w8-L2  & {\small25.31}                 & {\small25.61}                 & {\small48.09}                & {\small39.57}                   & {\small24.82}          & {\small15.77}                  & {\small23.33}                    & {\small\textbf{23.68}}                & {\small23.47}           & {\small70.00}      & {\small\textbf{86.17}}                   & {\small\textbf{42.76}}                 & {\small3.04}         & {\small79.96}              & {\small\textbf{54.58}}     & \multicolumn{1}{l|}{{\small\textbf{52.51}}}               & {\small$\:$\textbf{39.92}}       \\ 
                \small full-L2  & {\small\textbf{25.39}}                 & {\small\textbf{27.23}}                 & {\small44.24}                & {\small\textbf{43.47}}                   & {\small24.55}          & {\small\textbf{19.05}}                  & {\small\textbf{27.42}}                    & {\small22.94}                & {\small\textbf{26.14}}           & {\small\textbf{70.50}}      & {\small84.86}                   & {\small37.39}                 & {\small3.40}         & {\small54.84}              & {\small53.32}     & \multicolumn{1}{l|}{{\small48.84}}               & {\small$\:$38.35}       \\ \midrule
			
			\multicolumn{18}{c}{\small   C=512} \\
			\small H2O          & {\small21.72}                 & {\small26.03}                 & {\small44.81}                & {\small32.33}                   & {\small23.16}                   & {\small14.86}                  & {\small23.65}                    & {\small22.84}                & {\small24.70}                    & {\small42.00}               & {\small85.22}                   & {\small41.57}                 & {\small3.40}         & {\small86.45}              & {\small53.04}              & \multicolumn{1}{l|}{{\small49.68}}               & {\small$\:$37.22}                \\
			\small SnapKV       & {\small24.60}        & {\small27.81}                 & {\small48.98}       & {\small39.46}                   & {\small25.25}          & {\small16.98}                  & {\small23.70}                    & {\small22.96}                & {\small24.37}                    & {\small67.00}               & {\small85.88}                   & {\small41.26}                 & {\small2.78}                  & {\small86.56}              & {\small54.81}     & \multicolumn{1}{l|}{{\small51.71}}      & {\small$\:$40.26}                \\
			\small Pyramid      & {\small23.23}                 & {\small27.94}                 & {\small48.87}                & {\small40.50}                   & {\small24.36}                   & {\small16.74}                  & {\small23.22}                    & {\small23.16}                & {\small24.37}                    & {\small67.00}               & {\small85.73}                   & {\small41.74}                 & {\small3.16}                  & {\small85.67}              & {\small54.16}              & \multicolumn{1}{l|}{{\small50.34}}               & {\small$\:$40.01}                \\
			\small Ada-SKV   & {\small23.39}                 & {\small28.72}                 & {\small48.96}                & {\small40.60}          & {\small25.20}                   & {\small17.25}                  & {\small23.15}                    & {\small23.48}                & {\small24.41}                    & {\small68.00}      & {\small\textbf{86.39}}          & {\small41.69}                 & {\small2.73}                  & {\small\textbf{88.92}}     & {\small54.69}              & \multicolumn{1}{l|}{{\small51.51}}               & {\small$\:$40.57}       \\
			\small Ada-PKV  & {\small24.03}                 & {\small28.98}        & {\small48.39}                & {\small39.25}                   & {\small24.50}                   & {\small18.38}         & {\small23.13}                    & {\small23.90}       & {\small24.30}                    & {\small68.00}      & {\small85.89}                   & {\small41.89}        & {\small2.98}                  & {\small87.71}              & {\small54.46}              & \multicolumn{1}{l|}{{\small51.39}}               & {\small$\:$40.45}                \\ \midrule
                \small w32-L1  & {\small23.73}                 & {\small28.54}                 & {\small48.65}                & {\small40.42}                   & {\small26.33}          & {\small18.12}                  & {\small23.92}                    & {\small23.29}                & {\small24.65}           & {\small67.25}      & {\small86.09}                   & {\small41.86}                 & {\small2.81}         & {\small85.92}              & {\small54.55}     & \multicolumn{1}{l|}{{\small51.85}}               & {\small$\:$40.50}       \\ 
                \small w32-L2  & {\small24.52}                 & {\small28.59}                 & {\small\textbf{50.55}}                & {\small41.22}                   & {\small26.61}          & {\small18.35}                  & {\small24.49}                    & {\small\textbf{24.16}}                & {\small24.94}           & {\small70.00}      & {\small86.12}                   & {\small42.06}                 & {\small2.75}         & {\small88.43}              & {\small55.26}     & \multicolumn{1}{l|}{{\small52.92}}               & {\small$\:$\textbf{41.31}}       \\ 
                \small w8-L2  & {\small25.12}                 & {\small29.90}                 & {\small48.79}                & {\small39.79}                   & {\small\textbf{27.00}}          & {\small17.82}                  & {\small25.43}                    & {\small23.49}                & {\small25.20}           & {\small70.00}      & {\small86.15}                   & {\small\textbf{43.29}}                 & {\small2.42}         & {\small85.13}              & {\small\textbf{56.12}}     & \multicolumn{1}{l|}{{\small\textbf{53.35}}}               & {\small$\:$41.19}       \\ 
                \small full-L2  & {\small\textbf{25.23}}                 & {\small\textbf{29.97}}                 & {\small46.90}                & {\small\textbf{41.55}}                   & {\small25.60}          & {\small\textbf{19.59}}                  & {\small\textbf{29.25}}                    & {\small23.24}                & {\small\textbf{26.55}}           & {\small\textbf{70.50}}      & {\small85.91}                   & {\small39.50}                 & {\small\textbf{4.60}}         & {\small63.25}              & {\small54.04}     & \multicolumn{1}{l|}{{\small50.50}}               & {\small$\:$39.76}       \\ \midrule
			\multicolumn{18}{c}{\small   C=1024} \\
			\small H2O          & {\small23.90}                 & {\small28.62}                 & {\small46.46}                & {\small37.03}                   & {\small24.74}                   & {\small15.04}                  & {\small25.30}                    & {\small23.11}                & {\small25.92}                    & {\small46.00}               & {\small85.93}                   & {\small41.80}                 & {\small3.24}         & {\small86.57}              & {\small54.46}              & \multicolumn{1}{l|}{{\small51.01}}               & {\small$\:$38.70}                \\
			\small SnapKV       & {\small25.47}         & {\small29.57}                 & {\small\textbf{49.33}}       & {\small40.90}          & {\small25.53}                   & {\small19.01}                  & {\small25.94}                    & {\small23.89}                & {\small26.21}                    & {\small69.50}               & {\small\textbf{86.48}}          & {\small42.10}                 & {\small2.98}                  & {\small88.56}     & {\small55.57}     & \multicolumn{1}{l|}{{\small51.92}}               & {\small$\:$41.44}                \\
			\small Pyramid      & {\small24.21}                 & {\small29.86}                 & {\small48.93}                & {\small40.75}                   & {\small25.05}                   & {\small18.77}                  & {\small25.73}                    & {\small24.06}       & {\small25.65}                    & {\small68.50}               & {\small86.31}                   & {\small42.25}                 & {\small2.97}                  & {\small87.17}              & {\small54.75}              & \multicolumn{1}{l|}{{\small52.10}}               & {\small$\:$41.07}                \\
			\small Ada-SKV   & {\small24.79}                 & {\small\textbf{31.94}}        & {\small48.45}                & {\small40.73}                   & {\small26.22}                   & {\small19.11}         & {\small25.61}                    & {\small23.92}                & {\small26.03}                    & {\small70.00}      & {\small86.32}                   & {\small42.35}                 & {\small2.91}                  & {\small88.31}              & {\small55.44}              & \multicolumn{1}{l|}{{\small52.55}}      & {\small$\:$41.54}       \\
			\small Ada-PKV  & {\small25.09}                 & {\small30.94}                 & {\small48.18}                & {\small40.00}                   & {\small26.52}          & {\small19.10}                  & {\small24.93}                    & {\small23.71}                & {\small25.86}                    & {\small70.00}      & {\small86.34}                   & {\small42.64}        & {\small2.56}                  & {\small86.92}              & {\small54.93}              & \multicolumn{1}{l|}{{\small51.90}}               & {\small$\:$41.23}                \\ \midrule
                \small w32-L1  & {\small24.77}                 & {\small29.83}                 & {\small49.27}                & {\small41.52}                   & {\small26.09}          & {\small18.58}                  & {\small26.01}                    & {\small23.37}                & {\small25.99}           & {\small69.00}      & {\small86.22}                   & {\small42.49}                 & {\small3.19}         & {\small\textbf{89.81}}              & {\small55.61}     & \multicolumn{1}{l|}{{\small52.60}}               & {\small$\:$41.52}       \\ 
                \small w32-L2  & {\small25.43}                 & {\small31.89}                 & {\small49.32}                & {\small42.12}                   & {\small26.26}          & {\small18.44}                  & {\small27.00}                    & {\small23.84}                & {\small26.37}           & {\small70.50}      & {\small86.38}                   & {\small43.46}                 & {\small2.77}         & {\small88.31}              & {\small56.38}     & \multicolumn{1}{l|}{{\small52.64}}               & {\small$\:$41.94}       \\ 
                \small w8-L2  & {\small25.30}                 & {\small31.72}                 & {\small48.54}                & {\small\textbf{42.14}}                   & {\small\textbf{27.00}}          & {\small17.93}                  & {\small26.77}                    & {\small\textbf{24.24}}                & {\small26.00}           & {\small71.00}      & {\small86.10}                   & {\small\textbf{43.67}}                 & {\small2.61}         & {\small86.56}              & {\small\textbf{57.46}}     & \multicolumn{1}{l|}{{\small\textbf{54.09}}}               & {\small$\:$\textbf{41.95}}       \\ 
                \small full-L2  & {\small\textbf{26.16}}                 & {\small30.69}                 & {\small48.64}                & {\small41.80}                   & {\small26.27}          & {\small\textbf{19.58}}                  & {\small\textbf{31.41}}                    & {\small23.72}                & {\small\textbf{26.88}}           & {\small\textbf{71.00}}      & {\small85.68}                   & {\small41.87}                 & {\small\textbf{4.86}}         & {\small72.46}              & {\small55.56}     & \multicolumn{1}{l|}{{\small51.58}}               & {\small$\:$41.14}       \\ \bottomrule
		\end{tabular}
		\caption{Comparison of baseline methods against four KV-Compress variants for Mistral-7B-Instruct-v0.2 over 16 LongBench subsets. Results for H2O, SnapKV, PyramidKV (Pyramid), Ada-SnapKV (Ada-SKV) and Ada-PyramidKV (Ada-PKV) were taken from \cite{feng2024adakvoptimizingkvcache}. Highest score for each column shown in bold. For cache size of $C$, all baseline methods keep $C\times 32\times 32$ KVs in cache while our method keeps only $C\times 32\times \textbf{8}$ KVs. Table format taken from \cite{feng2024adakvoptimizingkvcache}.}
		\label{tab:detail_mistral}
\end{table*}
Table \ref{tab:detail_mistral} shows results on the LongBench suite with Mistral-7B-Instruct-v0.2. We benchmark four variations of KV-Compress against prior methods. We find L2 aggregation to demonstrate clear superiority over the L1 aggregation used in prior work and find that a smaller observation window of $w=8$ performs better that $w=32$ in most cases. We also evaluate \textit{KVC-full-L2}, a variation that uses full query range aggregation, similar to H2O, but with a small excluded query window. This variant actually performs the best in many subtasks, despite seeing large degradation for others, such as SAMSum, especially at high compression rates. Though the quadratic scaling of compute cost for running this algorithm makes it impractical for most cases, these results suggest that the effect of the range of queries used when scheduling key-value eviction should be studied more carefully.

We evaluate with Llama-3.1-8B-Instruct using the most performant of our methods, \textit{KVC-w8-L2}, which uses an observation window of size $w=8$ and uses an aggregation of squared attention when determining KV eviction. Results are shown in table \ref{tab:detail_llama3}. In this case we reach state-of-the-art results nearly across the board for the most extreme max-cache-size configuration, $C=128$. Indeed the only subsets we don't achieve the highest performance in are NarrativeQA, PassageCount and LCC.

We achieve state-of-the-art results for average LongBench performance across all max-cache-size configurations for both Mistral and Llama, with Mistral state-of-the-art being shared between \textit{w32-L2} and \textit{w8-L2} variants.

It should be noted that our state-of-the-art performance is achieved while using $\frac{1}{4}$ as many KVs as the baseline methods we test against. These baselines conduct compression after repeating KVs so that their shape matches the number of \textit{query} heads, while our methods compress over a non-repeated KV cache where shape is determined by the number of \textit{key-value} heads. This means that for the same maximum cache size our method holds $\frac{1}{r}$ as many KVs as the baselines, where $r$ is the ratio of query heads to key-value heads. For both Mistral-7B and Llama-3.1-8B, $r = 32 / 8 = 4$ so that our method achieves a 4x higher compression rate for equivalent maximum cache size configurations.

\subsection{Throughput Benchmarks}
Reducing KV cache memory enables larger batching, leading to improved total throughput in cases where available device memory is a limiting factor. This is commonly the case in single-instance LLM deployments, where most vRAM is allocated to model parameters.

To evaluate whether our method can improve throughput of a PagedAttention framework, we test our modification of vLLM against vanilla vLLM v0.6.0 across two single-instance configurations: Llama-3.1-8B-Instruct run on an NVIDIA L4, and Llama-3.1-70B-Instruct-FP8 run on an NVIDIA H100. Both configurations experience vRAM scarcity due to the large model size relative to global memory of the device used, and so are ideal for benchmarking our approach.

In Figure \ref{fig:llama-8b-throughput-thrpt} we plot total throughput by compression rate for Llama-3.1-8B on a single L4, over a range of input context lengths. We reach throughput multipliers of 4.93x and 5.18x over vanilla vLLM for compression rates of 32 and 64, respectively, for context length $L_c=6000$. For shorter contexts we continue to see significant improvement even at smaller compression rates, observing multipliers of 2x and 2.54x for compression rates of 2 and 4, respectively, with $L_c=500$.

Figure \ref{fig:llama-70b-throughput-thrpt} shows the same plot for an FP8 quantization of Llama-3.1-70B run on a single H100. The 70B model achieves a more reasonable throughput multipliers of 2.14x and 1.8x over vanilla vLLM, at compression rates of 64 and 8, respectively.

\subsection{LongBench with Continual Compression}
To determine the feasibility of conducting inference at the compression rates used in our throughput benchmarks, we measure performance on LongBench using the same configuration as our throughput benchmarks. We use a fixed compression rate and compress after every model iteration (including decoding steps) to test worst case performance when compression is allowed to occur anytime after prefill.

We evaluate performance of each subtask under each compression rate as a percentage of full-cache performance on that subtask. In figure \ref{fig:llama-8b-throughput-acc} we plot percent performance averaged over all subtasks in a category for Llama-3.1-8B-Instruct. It's immediately clear that summarizaton tasks are the most difficult, with performance degrading sharply even for the smallest compression rates. 
Full plots of each subtask's individual performance can be found in figure \ref{fig:llama-8b-longbench-all}

Category-average and individual subtask results for Llama-3.1-70B-Instruct-FP8 are found in figures \ref{fig:llama-70b-throughput-acc} and \ref{fig:llama-70b-longbench-all}, respectively.
We notice that the 70B model is much less sensitive to the compression than 8B. We see that nearly all non-summarization subtasks (aside from Qasper) mainain over 90\% performance even for a 64x compression rate (the maximum tested). however, that it is able to maintain its performance better than the 70B model, with a majority of tasks retaining over 95\% performance even at 64x compression. Even on the problematic Qasper dataset it is able to retain 90\% for all tested compression rates. Summarization tasks GovReport and QMSum remain difficult, however, with performance degrading quickly, even at low compression rates.

In past work, prompts for the LongBench coding subtasks have been templated to append the following line to the input context:
\begin{quote}
    \textit{Next line of code:}
\end{quote}
For Llama-3.1-70B-Instruct-FP8, we find this confuses the model, leading to lower performance in the full-cache case. Interestingly, we find that KV cache compression resolves this issue, yielding \textit{improvement} in overall performance. Still, to provide the strongest full-cache baseline possible, we remove this appended line from the templates of subtasks in the coding category for all experiments with the 70B model.

\subsection{Analysis of Decoding Batch Size}
\begin{figure}
\centering
\begin{subfigure}{.5\textwidth}
  \centering
  \caption{ }
  \vspace{-30pt}
  \includegraphics[width=1.05\linewidth]{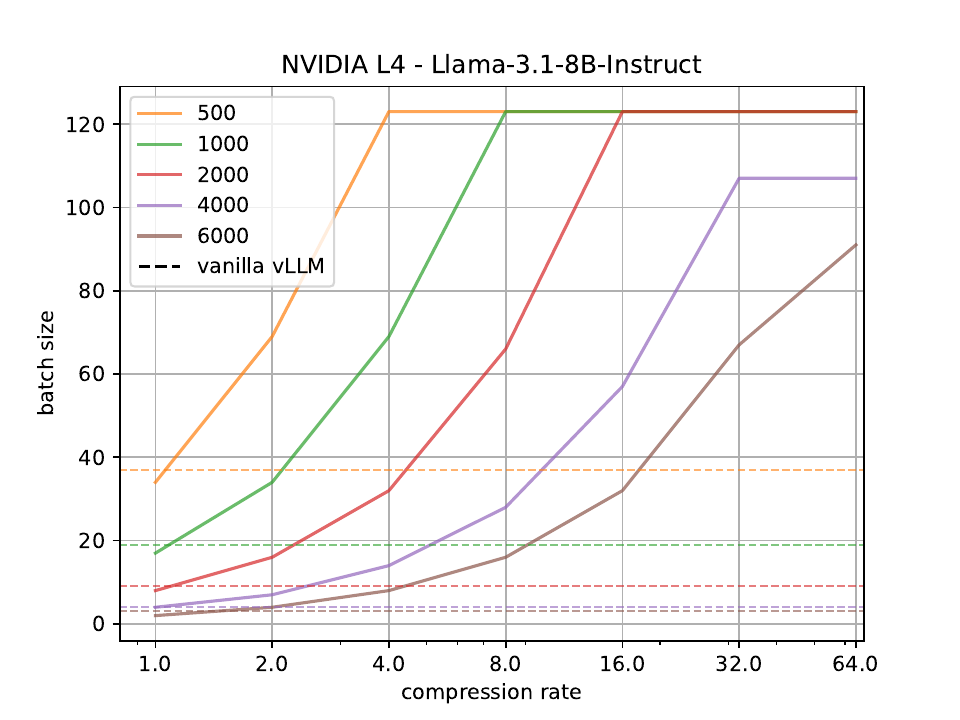}
  \label{fig:max-batch-size-llama-3.1-8b}
\end{subfigure}%
\begin{subfigure}{.5\textwidth}
  \centering
  \caption{ }
  \vspace{-30pt}
  \includegraphics[width=1.05\linewidth]{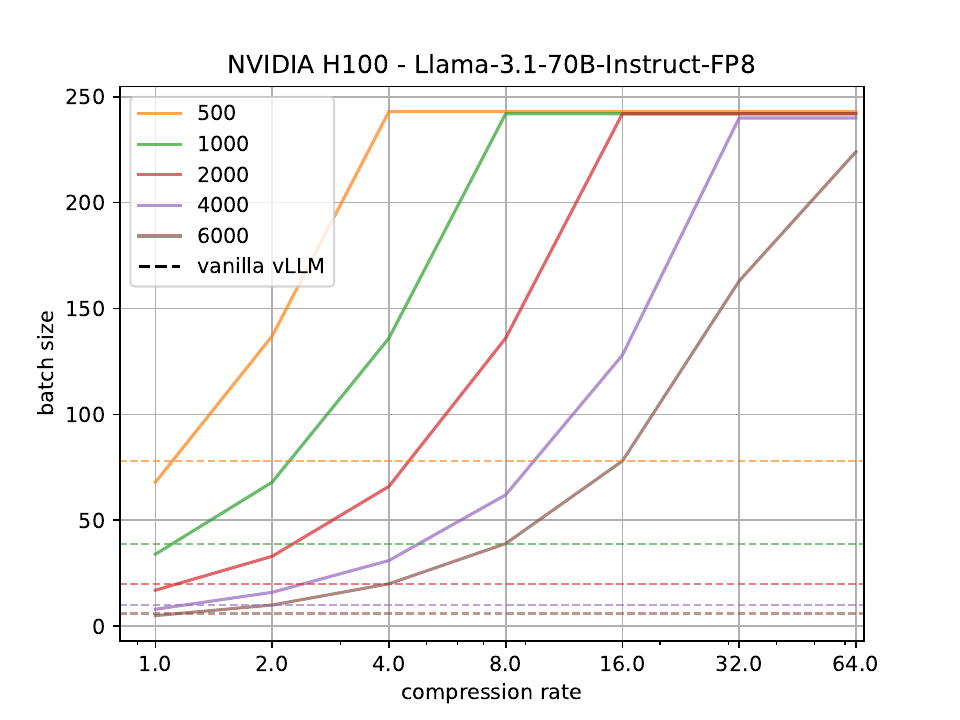}
  \label{fig:max-batch-size-llama-3.1-70b}
\end{subfigure}
\caption{Maximum decoding batch size by compression rate. Plots shown for multiple input token length configurations.}
\end{figure}

KV-Compress is able to boost throughput of vLLM by increasing the availability of free cache blocks and allowing more sequences to be loaded into the decoding batch at once and decoded in parallel.

To study the effect of our compression on batch size more directly, we plot by compression rate the maximum decoding batch size that was reached during each benchmarking run in figures \ref{fig:max-batch-size-llama-3.1-8b} and \ref{fig:max-batch-size-llama-3.1-70b} for Llama-3.1 8B and 70B, respectively. We again visualize across multiple input context lengths. While one might expect a strictly linear relationship between maximum batch size and compression rate, we note that larger input context lengths require a larger compression rate before near-proportional increases in batch size are observed. This is because our integration requires that a sequence be prefilled before being compressed. Even if we have room in cache for 10 more \textit{compressed} sequences, if we cannot load one more \textit{uncompressed} sequences into cache then we will still be unable to add to our batch. Having a distribution of input context lengths in the model's requests (as will be the case in practice) should alleviate this issue to some extent. 

Nonetheless, we are able to reach batch sizes of over 100 for both configurations in spite of the limited memory available. 

\subsection{Effect of Context Length on Throughput}
\begin{figure}
\centering
\begin{subfigure}{.5\textwidth}
  \centering
  \caption{ }
  \vspace{-30pt}
  \includegraphics[width=1.05\linewidth]{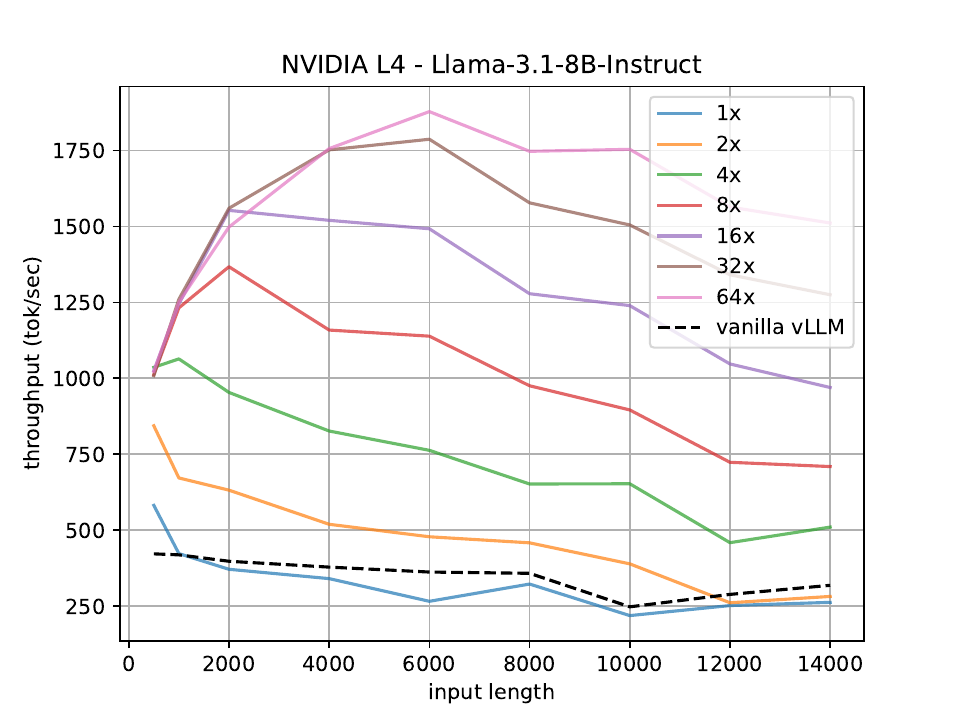}
  \label{fig:thrpt-by-len-llama-3.1-8b}
\end{subfigure}%
\begin{subfigure}{.5\textwidth}
  \centering
  \caption{ }
  \vspace{-30pt}
  \includegraphics[width=1.05\linewidth]{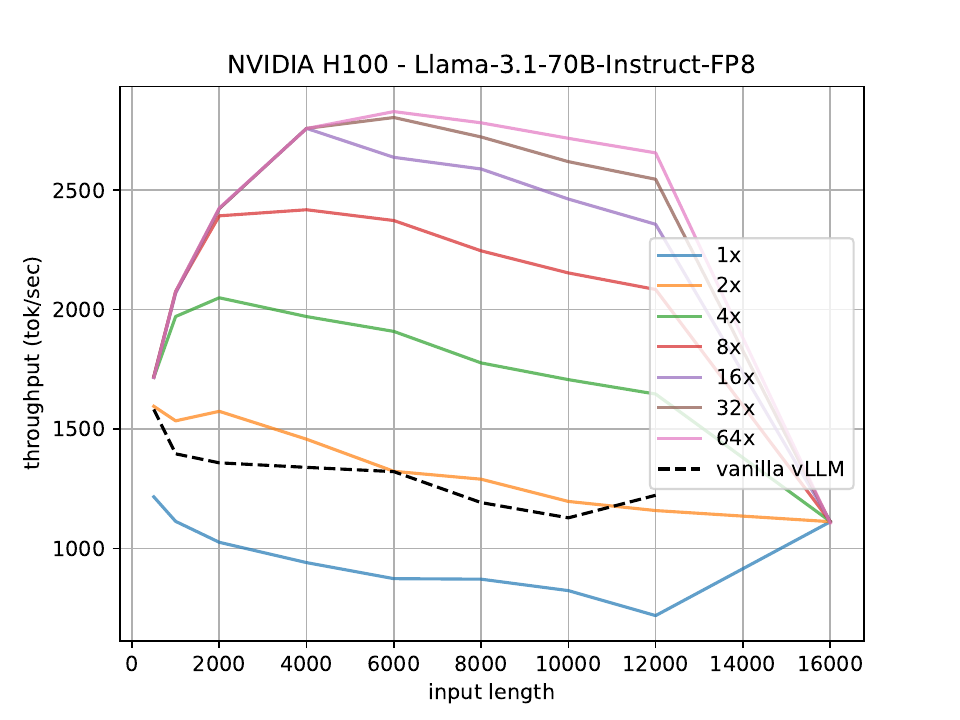}
  \label{fig:thrpt-by-len-llama-3.1-70b}
\end{subfigure}
\caption{Total throughput by input token length. Plots shown for compression rates from 1-64x.}
\end{figure}
To better understand the interplay between context length, compression rate and throughput, in figures \ref{fig:thrpt-by-len-llama-3.1-8b} and \ref{fig:thrpt-by-len-llama-3.1-70b} we plot throughput of both models by context length for several compression rate configurations side-by-side.

We plot over input context lengths $L_c=\{500, 1000, 2000, 4000, 6000, 8000, 10000, 12000\}$ to observe how throughput behaves as input context approaches the limit of allocable cache tokens. We observe that the relative throughput under compression begins decreasing as new input context length approaches this limit and new sequences must wait longer before being added to the decoding batch.

\section{Conclusion}

We have presented KV-Compress, a framework for compression of a paged KV cache. With our modifications to prior work, including aggregating squared attention, allowing variable eviction rates across both layers and heads, and evicting from a non-repeated KV cache, KV-Compress leads by a comfortable margin on the established benchmarks for cache compression with Mistral-7B-Instruct-v0.2 and Llama-3.1-8B-Instruct. We have shown that our method can reach compression rates of 8-64x without significantly impacting model performance, and that, with such compression rates, we can boost throughput of state-of-the-art inference frameworks many times over.

\bibliographystyle{unsrt}  
\bibliography{references}

\end{document}